\documentclass[journal]{IEEEtran}
\usepackage{amsmath,amsfonts}
\usepackage{algorithmic}
\usepackage{algorithm}
\usepackage{array}
\usepackage[caption=false,font=small,labelfont=sf,textfont=sf]{subfig}
\usepackage{textcomp}
\usepackage{stfloats}
\usepackage{url}
\usepackage{verbatim}
\usepackage{graphicx}
\usepackage{cite}

\usepackage{siunitx}

\usepackage{multirow}

\usepackage{makecell}
\usepackage{booktabs}
\usepackage{hhline} 

\usepackage{amsthm}

\usepackage[hidelinks]{hyperref}

\usepackage{float}
\newtheorem{lemma}{Lemma}
\newtheorem{theorem}{Theorem}

\newcommand{\ii}[0]{I\hspace{-0.3mm}I}
\newcommand{\iii}[0]{I\hspace{-0.3mm}I\hspace{-0.3mm}I}

\begin{document}

\title{RGBlimp-Q: Robotic Gliding Blimp\\ With Moving Mass Control \\Based on a Bird-Inspired Continuum Arm}

\author{Hao Cheng and Feitian Zhang%
\thanks{This research was in part supported by the National Natural Science Foundation of China under Grant 62473006. 
\textit{(Corresponding author: Feitian Zhang.)}}%
\thanks{The authors are with the Robotics and Control Laboratory, School of Advanced Manufacturing and Robotics, and the State Key Laboratory of Turbulence and Complex Systems, Peking University, Beijing, 100871, China (e-mail: h-cheng@stu.pku.edu.cn; feitian@pku.edu.cn).}%
\thanks{The videos, open-sourced hardware, and source code of this work are available at https://RGBlimp.github.io.}%
}

\maketitle

\begin{abstract}
Robotic blimps, as lighter-than-air aerial platforms, offer extended operational duration and enhanced safety in \mbox{human}-robot interactions due to their buoyant lift. 
However, achieving robust flight performance under environmental airflow disturbances remains a critical challenge, thereby limiting their broader deployment. 
Inspired by avian flight mechanics, particularly the ability of birds to perch and stabilize in turbulent wind conditions, this article introduces \mbox{RGBlimp-Q}---a robotic gliding blimp equipped with a bird-inspired continuum arm featuring a novel moving mass actuation mechanism. 
This continuum arm enables flexible attitude regulation through internal mass redistribution, significantly enhancing the system's resilience to external disturbances. 
Additionally, it facilitates aerial manipulation by employing end-effector claws that interact with the environment in a manner analogous to avian perching behavior. 
This article presents the design, modeling, and prototyping of \mbox{RGBlimp-Q}, supported by comprehensive experimental evaluation and comparative analysis. 
To the best of the authors' knowledge, this represents the first interdisciplinary integration of continuum mechanisms into a lighter-than-air robotic platform, where the continuum arm simultaneously functions as both an actuation and manipulation module. This design establishes a novel paradigm for robotic blimps, expanding their applicability to complex and dynamic environments. 
\end{abstract}


\section{Introduction}
\label{sec.intro}
\IEEEPARstart{W}{ith} the growing demand for aerial vehicle applications, there is an increasing need for portable miniature aerial robots capable of low-speed, safe, and long-endurance flight \cite{ManipSurvey,ManipSurvey2,ManipSurvey3}. 
In recent years, significant attention has been directed towards the development of robotic blimps, owing to their extended flight endurance and enhanced safety during human-robot interactions \cite{Blimp7,Blimp6,searchRescue1,RGBlimp,Blimp8,Blimp4,Blimp10,xu2023sblimp}. 
As a type of lighter-than-air (LTA) aerial vehicle, robotic blimps offer substantial potential for applications that require prolonged airborne operation at low speeds. 
Examples include search and rescue missions \cite{searchRescue1,searchRescue2}, environmental monitoring \cite{EnvMoni1,EnvMoni2}, human-robot interaction \cite{HRI0,HRI1,HRI2}, and entertainment \cite{Joy1,Joy2}. 
Unlike heavier-than-air (HTA) aircraft, robotic blimps leverage buoyancy to counteract gravity, resulting in reduced energy consumption. 
However, their buoyant nature makes them particularly sensitive to airflow disturbances, which poses a significant challenge to achieving robust flight, especially in complex outdoor environments. 
Therefore, enhancing the flight stability and robustness of robotic blimps is crucial to fully realizing their potential for long endurance and safe operation. 

\begin{figure}[t]
      \centering
      \includegraphics[width=86mm]{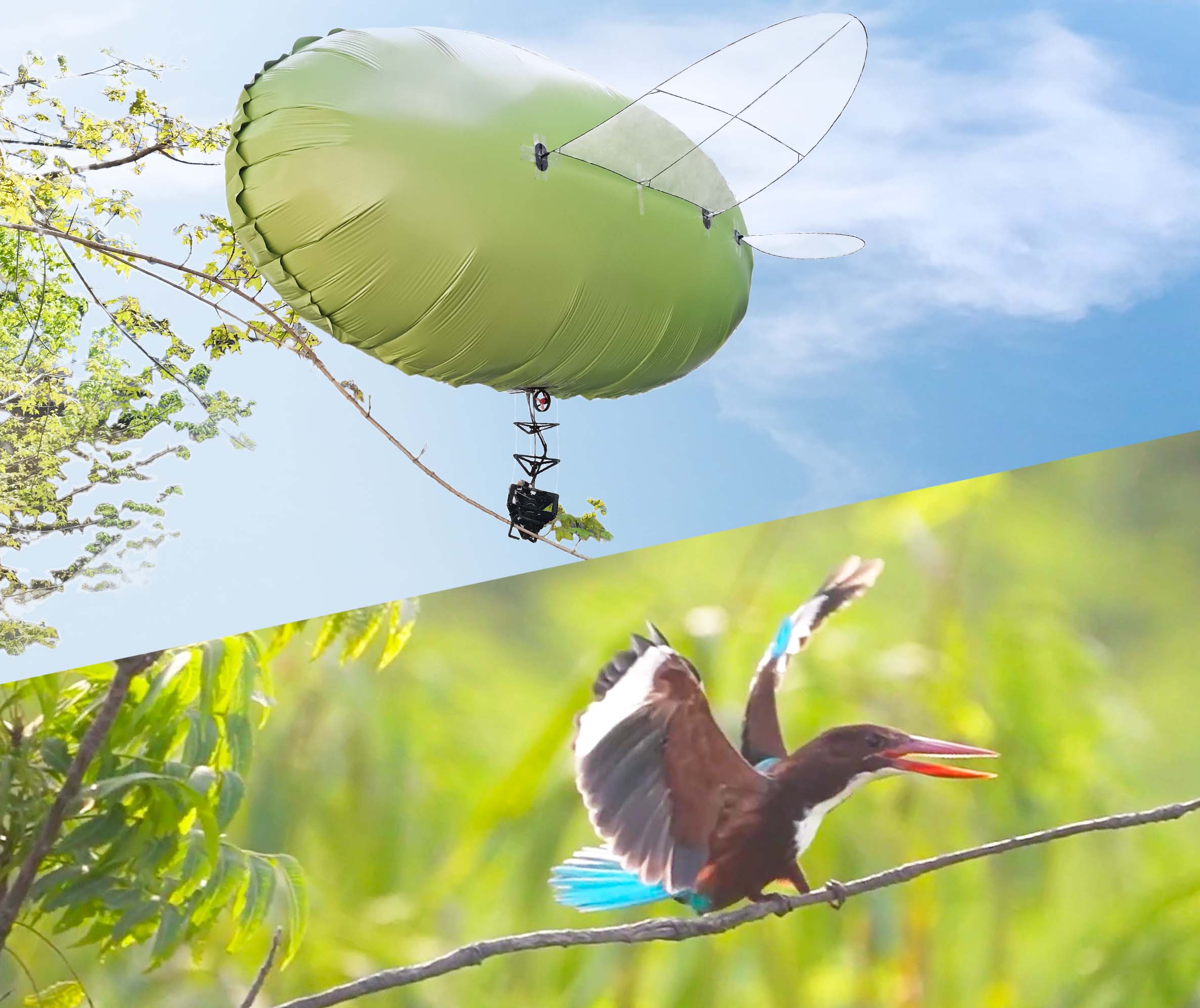}
      \caption{A snapshot of the \mbox{RGBlimp-Q} perching on a branch through the use of the novel continuum arm design (the upper figure), emulating the behavior of a bird, as shown in the lower figure \cite{birdPic1}.} 
      \label{fig.1}
\end{figure} 

\IEEEpubidadjcol

Numerous efforts in the literature have focused on achieving robust flight for robotic blimps by employing disturbance compensation controllers while maintaining the conventional omnidirectional thrust actuation mechanism \cite{robustBlimp1a,robustBlimp1c,robustBlimp2,robustBlimp3a}, a method commonly used in rotary-wing aerial robots \cite{DroneRobust1,DroneRobust2,DroneRobust3,DroneRobust4}. 
With the advancement of rotary-wing aerial robots, significant progress has been made in the robust flight control of small-sized aerial vehicles. 
Notable examples include neural network-based disturbance estimators for robust flight control of quadrotors \cite{DroneRobust1}, uncertainty-aware model predictive controllers for mitigating wind and payload disturbances in multi-rotor systems \cite{DroneRobust2,DroneRobust3}, and controllers compensating for aerodynamic disturbances caused by ground effects \cite{DroneRobust4}. 
Building on methodologies from rotary-wing systems, disturbance compensation strategies for indoor robotic blimps have been demonstrated. 
These include Luenberger observer design for disturbance estimation \cite{robustBlimp1a}, wind effect mitigation via reinforcement learning \cite{robustBlimp2}, wind-aware motion planning using dynamic programming \cite{robustBlimp3a}, and attitude stabilization with dual-loop control \cite{Swing}. 
However, significant performance distinctions between robotic blimps and rotary-wing aerial robots, such as differences in aerodynamics and actuation mechanisms, have been underexplored. 
Consequently, (almost) all robust controllers based on the existing mechanism design of robotic blimps face challenges in achieving the robust outdoor flight performance seen in rotary-wing aerial robots. 
The mechanism design, particularly the actuation mechanism, specifically tailored for robotic blimps to enable robust flight has long been neglected. This oversight has hindered the successful application of robust flight control techniques to robotic blimps, preventing them from achieving the outdoor flight capabilities that rotary-wing aerial robots have demonstrated in windy environments. 
To address this challenge, it is essential to design a novel actuation mechanism for robotic blimps that enables robust flight performance. 
\begin{figure*}[!t]
      \centering
      \includegraphics[width=1\textwidth]{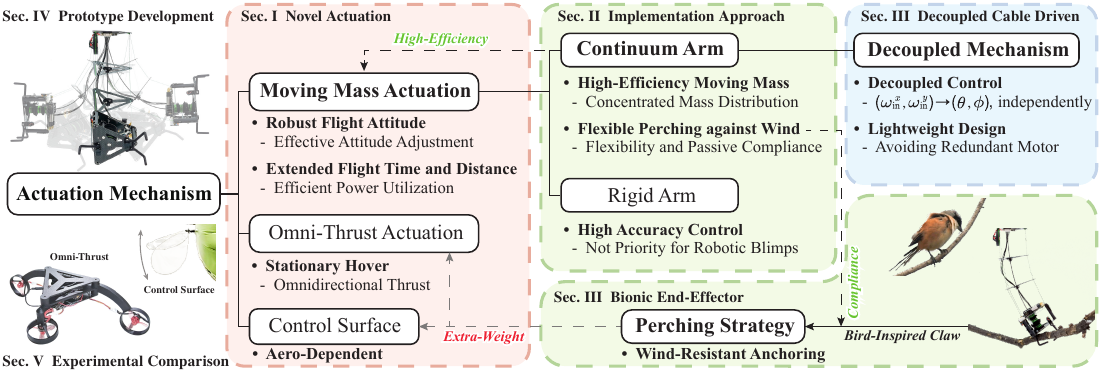}
      \caption{The hierarchical block diagram illustrating the structure of this article and the contributions of the proposed \mbox{RGBlimp-Q} design. 
       Section~\ref{sec.intro} introduces the moving mass actuation mechanism for \mbox{RGBlimp-Q}, which enables both effective attitude adjustment and efficient energy utilization. 
       This mechanism distinguishes \mbox{RGBlimp-Q} from conventional blimps, supporting sustained flight over extended time and distance for tasks such as indoor inspection and outdoor monitoring. 
       Section~\ref{Sec.model} proposes the continuum mechanism as an efficient implementation of moving mass actuation. 
       This mechanism is seamlessly integrated into the LTA aerial system to enhance moving mass actuation and enable perching capability through the end-effect claw. 
       Section~\ref{Sec.continuum} presents a lightweight, decoupled cable-driven mechanism designed to accommodate the limited payload capacity of robotic blimps. 
       Section~\ref{Sec.prototype} showcases the developed \mbox{RGBlimp-Q} prototype, equipped with the continuum arm. 
       Section~\ref{Sec.experiment} outlines indoor and outdoor flight experiments that validate the proposed design's effectiveness in attitude adjustment and energy utilization, demonstrating superior performance compared to conventional omnidirectional thrust mechanisms. 
       Furthermore, the reliability and repeatability of the continuum arm were confirmed through an 18-hour endurance test involving 3,000 repeated flight cycles. 
       } 
      \label{fig.2} 
\end{figure*}

In the natural world, flying lizards and flying squirrels utilize the inertia of their tails to control their gliding attitude, particularly during low-speed flight \cite{FlyingLizards,FlyingSquirrel}.
Similarly, birds exhibit robust flight by dynamically adjusting their wing movements to alter their attitude in response to wind disturbances \cite{Bird,Bird2} and by using their claws to perch on branches, effectively resisting excessive wind \cite{Perch1,birdPic2}. 
In contrast, current robotic blimps lack an adaptive mechanism analogous to these natural strategies, limiting their ability to withstand wind disturbances. 

With the rise of bio-inspired robotics \cite{Bioinspired} and aerial robotic manipulation \cite{ManipSurvey,ManipSurvey2}, significant progress has been made in the perching capabilities of rotary-wing aircraft. 
This progress focuses on enabling perching on support structures using manipulators, including bio-inspired passive mechanisms and actively controlled grippers \cite{Perch1,Perch2,Perch3,Perch4}. 
Examples include modular quadrotor landing gear \cite{Perch2}, bio-inspired robotic legs with grippers for rotary-wing drone perching \cite{Perch1}, and eagle-like claws for object capture in fixed-wing robots \cite{Manip1}. 

While rigid-arm-based aerial robotic manipulations remain dominant in the literature, continuum arms present a promising alternative due to their compact structure, lightweight design, and agile motion capabilities.  
This innovative concept incorporates flexible backbone structures with theoretically infinite degrees of freedom \cite{ContinuumSurvey,ContinuumSurvey2,ContinuumSurvey3}, utilizing various backbone materials such as elastic materials \cite{Continuum1,Continuum2,Continuum3,Continuum4,ContinuumAPCC}, and actuated by pneumatic or cable-driven mechanisms \cite{Continuum2,Continuum4,ContinuumAPCC,Continuum3}. 
To manage the inherent complexity of continuum arms, the piecewise constant curvature (PCC) kinematics is commonly employed \cite{ContinuumPCC,QParam,ContinuumAPCC,ContinuumIMU}. 
Recent advancements have further integrated aerial continuum manipulators onto quadrotors to facilitate grasping operations \cite{ManipContinuum1,peng2025dexterous}. 

However, the integration of such mechanisms onto robotic blimps remains largely unexplored. 
Given the inherent advantages of long endurance and low-speed flight, as well as the challenges posed by vulnerability to disturbances, the introduction of such mechanisms for aerial manipulation or perching in wind is both appropriate and essential for enhancing the functionality of LTA aerial systems. 
A significant challenge therein lies in the mismatch between the weight of existing manipulators and the limited payload capacity of robotic blimps. 

This article introduces \mbox{RGBlimp-Q}, \textbf{the first} robotic gliding blimp equipped with a bionic, lightweight continuum arm designed for both moving mass control and object capture. 
Building upon our previous work on RGBlimp \cite{RGBlimp}, which utilized hybrid aerodynamic-buoyant lift, \mbox{RGBlimp-Q} integrates a lightweight continuum arm with a bird-like claw.  
This bird-inspired continuum arm not only enables flexible spatial moving mass control, thereby enhancing effective attitude adjustment in the air, but also facilitates object grasping, mimicking the behavior of birds.
This design addresses a critical yet often overlooked requirement for station-keeping operations and wind-resistant anchoring in robotic blimps. 
A prototype of \mbox{RGBlimp-Q} is developed, weighing approximately $\SI{7}{g}$, accounting for net buoyancy. 
Extensive flight experiments, conducted both indoors and outdoors using a classical feedback controller, validate the effectiveness of the proposed design in improving flight maneuverability and robustness. 
As illustrated in Fig.\!~\ref{fig.2}, the contributions of this article are twofold: 

\textit{1) Novel actuation mechanism}: 
This article introduces the \mbox{RGBlimp-Q} design, a novel robotic blimp featuring a unique moving mass actuation mechanism for attitude adjustment. 
The proposed design aims to extend the advantages of robotic blimps to complex, disturbed environments. 
It enables both effective attitude control and efficient energy utilization, facilitating robust and durable flight over time and distance. 

\textit{2) Bird-inspired continuum arm}: 
The proposed bird-inspired continuum arm enhances moving mass actuation through concentrated mass distribution and enables object capture via the end-effector claw. 
This allows the robot to maintain stable flight or perch against disturbances, mimicking bird behavior. 
Additionally, a decoupled cable-driven mechanism is introduced to achieve a lightweight design and decoupled control in robotic blimps. 
This integration marks the first organic combination of a continuum mechanism and a cable-driven system in robotic blimp platforms. 

The aforementioned contributions are validated through systematic indoor and outdoor physical flight experiments, which demonstrate enhanced robustness in flight attitude, extended flight time and distance, and improved adaptability to complex environments with disturbances. 

The rest of this article is organized as follows. 
\mbox{Section}~\ref{Sec.model} presents the \mbox{RGBlimp-Q} design with improved flight attitude adjustment capability through moving mass actuation. 
Such novel actuation, implemented via a continuum arm, marks the first integration into an LTA system. 
Section~\ref{Sec.continuum} proposes the lightweight-oriented decoupled cable-driven continuum arm with an end-effect claw for both effective attitude control and resisting wind disturbances. 
Section~\ref{Sec.prototype} presents the developed \mbox{RGBlimp-Q} prototype and provides system identification results. 
Section~\ref{Sec.experiment} describes indoor and outdoor flight experiments that validate the effectiveness of the proposed design in both attitude adjustment and energy utilization, outperforming conventional omnidirectional thrust mechanisms. 
Moreover, the reliability and repeatability of the proposed continuum arm are validated through a repeated motion test exceeding $\SI{3000}{cycles}$ and $\SI{18}{hours}$. 
Finally, Section~\ref{Sec.conclusion} concludes this article and discusses potential improvements.

\section{RGBlimp-Q With Moving Mass Actuation}
\label{Sec.model}
To enable low-speed, long-duration flight in both indoor and outdoor environments while improving robustness against wind disturbances, we propose RGBlimp-Q, a robotic gliding blimp actuated via moving mass control. 
Our prior work on RGBlimp \cite{RGBlimp} demonstrated the critical role of aerodynamic optimization, such as the integration of wings and tail-planes, in enhancing the performance of robotic blimps. 
These enhancements were validated through aerodynamic analysis and empirical evaluation. 
Building upon these findings, achieving further aerodynamic robustness necessitates enhanced attitude adjustment capabilities. This is due to the sensitivity of flight dynamics to aerodynamic angles, specifically the angle of attack ($\alpha$) and sideslip angle ($\beta$), both of which are affected by the vehicle's attitude and environmental wind conditions (wind direction and strength). 
This section investigates the advantages of the moving mass control approach, focusing on its impact on flight maneuverability and robustness.  
It begins with an introduction to the moving mass actuation mechanism, highlighting the benefits of using the continuum mechanism for moving mass implementation. The section then presents the first-principle dynamic model of the \mbox{RGBlimp-Q}, which incorporates continuum-based moving mass actuation. 
To ensure generalization, the dynamics modeling is structured hierarchically, progressing from the general motion dynamics that account for moving mass actuation (Section~\ref{Sec.motion}) to continuum-specific mechanics (Section~\ref{Sec.ContinnumModel} and Section~\ref{Sec.full}), and concluding with the decoupled cable-driven control (\mbox{Section}~\ref{Sec.CableDriven} and \mbox{Section}~\ref{Sec.Control}).

\subsection{Moving Mass Actuation for Flight Attitude Adjustment} 
The moving mass actuation mechanism has historically been overlooked in the design of robotic blimps.
Previous design of robotic blimps in the literature often treated the non-negligible distance between the center of gravity (CG) and the center of buoyancy (CB) as a drawback and sought to mitigate its effects, such as swing-reducing control \cite{Swing}. 
Emerging research has shown the advantages of this inherent property. 
For example, Xu \textit{et\,\,al.} systematically utilize the inherent CG-CB distance to generate stabilizing swing moments for translational control \cite{xu2023sblimp}. 
In contrast, we hypothesize that transforming this fixed distance into a variable facilitates attitude adjustment and is achievable through moving mass control. 

This article proposes a unique robotic blimp design based on the moving mass actuation mechanism for improved attitude adjustment capability, distinguishing it from existing robotic blimps utilizing omnidirectional thrusters. 
This design leverages the gravitational moment associated with the variable spatial distance between CG and CB to effectively control the attitude of a robotic blimp, especially during low-speed flight. 
Conventional control surfaces, such as the rudder, elevator, and aileron, commonly used in high-speed aerial vehicles such as fixed-wing aircraft, exhibit significantly reduced effectiveness in controlling low-speed robotic blimps. 
This limitation comes from the dependence of aerodynamic effects on dynamic pressure, e.g., $\left \| \boldsymbol{M} \right \| \propto \rho V^2$, where $\boldsymbol{M}$ represents the aerodynamic moment, $\rho$ represents air density, and $V$ stands for the relative airspeed, equivalent to the cruising speed in calm environmental conditions. 

\begin{figure}[hbt]
      \centering
      \includegraphics[width=88mm]{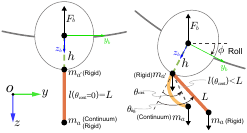} 
      \caption{Illustration of reference frames, forces, and key variables for theoretical comparison of roll adjustment via moving mass actuation: continuum arm (compact with concentrated mass $m_a$) versus rigid arm (with $m_a$ and joint actuator mass $m_\mathrm{a'}$). 
         Initially, both arms have zero rotation angle, i.e., $\theta_\mathrm{cont}\!=\!\theta_\mathrm{rig}\!=\!0$, resulting in a roll angle $\phi\!=\!0$ (left). 
         Gravity balance is achieved by moment equilibrium, i.e., $\tau_{m_{a'}}\!+\!\tau_{m_{a}}\!=\!0$ for the rigid arm, while $\tau_{m_{a}}\!=\!0$ for the continuum arm (right).}
      \label{fig.3}
\end{figure}

To achieve moving mass control, we employ a fixed-length continuum arm with two degrees of freedom (DoF), operating under the constant curvature (CC) assumption. 
This continuum mechanism, characterized by a compact mass distribution, enhances the attitude adjustment capabilities within the context of moving mass actuation for robotic blimps. 
It is identified as a crucial design choice for the \mbox{RGBlimp-Q}, and detailed in Section~\ref{Sec.ContinnumModel}. 
To illustrate, we examine roll angle $\phi\!\in\!\left(\SI{-90}{^\circ}\!, \SI{90}{^\circ}\right)$ adjustment, which is applicable to pitch as well, through a comparative study of moving mass actuation using continuum versus rigid mechanisms, as shown in Fig.\!~\ref{fig.3}. 

The comparative analysis is based on an idealized model wherein the blimp's center of gravity (CG) coincides with its center of buoyancy (CB), thereby isolating the manipulator's influence on moment generation. 
Under this assumption, we compare actuation efficiency between a continuum arm modeled as a single moving mass $m_a$ at its end and a rigid arm comprising the same $m_a$ plus an additional joint actuator mass $m_{\mathrm{a'}}$. 
Both arms are mounted at a distance $h$ below CG (CB)  and have the same constant length $L$. 
To ensure physically meaningful comparisons, the rotational angles $\theta_{\mathrm{cont}}$ and $\theta_{\mathrm{rig}}$, as actuation inputs, are constrained within the operational range of $[-60^\circ, 60^\circ]$.  

The constraint equation for the rigid arm, considering the equilibrium of gravitational moments, is derived as follows 
\begin{equation}
   \cot\phi_\mathrm{rig} = -\cot\theta_\mathrm{rig} - (1\!+\!k_m^{\mathrm{rig}})k_l^{\mathrm{rig}}\csc\theta_\mathrm{rig},
    \label{eq.1}
\end{equation}
where the coefficient $k_m^{\mathrm{rig}}\!=\!m_{a'}/m_a\!\geq\!0$ and $k_l^{\mathrm{rig}}\!=\!h/L\!>\!0$. 

In contrast, the CC assumption and the concentrated mass distribution of the continuum arm imply that its equilibrium is governed by geometric relations. 
Consequently, the constraint equation for the continuum arm related to the arc chord length $l(\theta_\mathrm{cont})$ is given by
\begin{equation}
    \cot\phi_\mathrm{cont} = -\cot\theta_\mathrm{cont} - k_l^{\mathrm{cont}}\theta_\mathrm{cont}\csc^2\theta_\mathrm{cont},
    \label{eq.2}
\end{equation}
where the coefficient $k_l^{\mathrm{cont}}\!=\!h/l(\theta_\mathrm{cont})\!>\!0$. 

Expanding Eqs.\!~\eqref{eq.1} and \eqref{eq.2}, the first-order approximation of these constraint equations show a negligible error of approximately $\pm 3^\circ$, computed as 
\begin{equation}
    \displaystyle
    \phi_\mathrm{cont} = -K_\mathrm{cont}\theta_\mathrm{cont},\qquad
    \phi_\mathrm{rig}  = -K_\mathrm{rig}\theta_\mathrm{rig},
    \label{eq.3}
\end{equation}
Here, the coefficients $K_\mathrm{cont}$ and $K_\mathrm{rig}$ represent the scaling factors for the continuum and rigid arms, respectively. 
These factors quantify the extent to which the rotational angles of the arms influence the roll angle, calculated as
\begin{equation}
    \textstyle K_\mathrm{cont}=\tfrac{1}{(1+k^{\mathrm{cont}}_l)},\qquad
    K_\mathrm{rig}=\tfrac{1}{1+(1+k^{\mathrm{rig}}_m)k_l^{\mathrm{rig}}}.\nonumber
\end{equation}

We compare the attitude adjustment capabilities by analyzing $K_\mathrm{cont}$ and $K_\mathrm{rig}$ under the condition that $\theta_\mathrm{cont}\!=\!\theta_\mathrm{rig}$. 
Given the same arm length $L$ and displacement $h$ in Eq.\!~\eqref{eq.3}, the continuum arm exhibits a higher gain due to its concentrated mass distribution. 
Specifically, $K_\mathrm{cont}$ is independent of $k_m$, leading to a greater gain $K_\mathrm{cont}$ compared to $K_\mathrm{rig}$, as $k^{\mathrm{rig}}_m\!=\!m_{a'}/{m_a}>0$ for rigid arms. 
For example, with parameters $L\!=\!\SI{0.40}{m}$, $h\!=\!\SI{0.30}{m}$, $m_a\!=\!\SI{30}{g}$, and $m_{a'}\!=\!\SI{15}{g}$, $K_\mathrm{cont}\!\approx\!0.57, K_\mathrm{rig}\!\approx\!0.47$, indicating an approximately $\SI{21.3}{\%}$ improvement in attitude adjustment achieved by the continuum actuation mechanism. 
For maximum input rotation angles $\theta_\mathrm{cont}\!=\!\theta_\mathrm{rig}\!=\!\pm\SI{60}{^\circ}$, the resulting roll angles are  $|\phi_\mathrm{cont}|\!=\!\SI{34.2}{^\circ}$ and $|\phi_\mathrm{rig}|\!=\!\SI{28.2}{^\circ}$, respectively. 
Furthermore, for the same attitude angle $\phi_\mathrm{cont}\!=\!\phi_\mathrm{rig}$ (as Fig.\!~\ref{fig.3}), the continuum arm requires only $K_\mathrm{rig}/K_\mathrm{cont}\!\approx\!\SI{82}{\%}$ of the input angle needed for the rigid arm. 
In conclusion, the continuum mechanism, with its compact mass distribution, provides superior attitude adjustment efficiency within the moving mass actuation framework, thereby reinforcing its pivotal role in the \mbox{RGBlimp-Q} design. 
Note that while the designed continuum arm shows higher compliance, further analysis is needed for generalized conclusions. 

\subsection{Motion Dynamics Modeling of RGBlimp-Q}
\label{Sec.motion}
\begin{figure}[ht]
      \centering
      \includegraphics[scale=1]{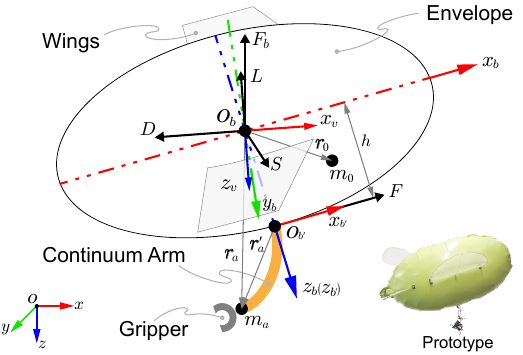} 
      \caption{Illustration of the \mbox{RGBlimp-Q} model, featuring a buoyant envelope, a continuum arm, and a pair of wings. The mass distribution includes the stationary mass $m_0$ and the internal movable mass $m_a$, with displacements $r_0$ and $r_a$ relative to the center of buoyancy (CB), respectively. The continuum arm is equipped with a gripper, serving as an aerial robotic manipulator, as discussed in Section~\ref{Sec.BioClaw}. 
      A snapshot of the \mbox{RGBlimp-Q} prototype is shown in the lower-right corner. 
      }
      \label{fig.4}
\end{figure}

In this section, we derive the translational and rotational dynamics model of the \mbox{RGBlimp-Q}, considering the continuum arm-based moving mass control. 
The \mbox{RGBlimp-Q} is modeled as a rigid-body system, subject to inertial forces and moments exerted by the moving mass within the robot, as well as aerodynamic forces and moments arising from the surrounding airflow, as illustrated in Fig.\!~\ref{fig.4}. 
The actuation of the system comprises a continuum arm serving as an internal movable mass $m_a$ for moving mass control, along with a propeller generating forward thrust $F$. 

The dynamics model is conceptualized as a system of two point masses, as shown in Fig.\!~\ref{fig.4}. 
The stationary mass, including wings and an envelope filled with helium, is denoted as $m_0$, characterized by a constant displacement $\boldsymbol{r}_0$ with respect to CB. 
The internal movable mass, represented by a continuum arm with a gripper, is denoted as $m_a$, featuring a controllable displacement $\boldsymbol{r}_a$ with respect to CB. 

The coordinate frames and external forces are defined following the convention \cite{FlightDynamic} and illustrated in Fig.\!~\ref{fig.4}. 
The position and orientation of the robot are defined with respect to the earth-fixed inertial frame $O\text{-}xyz$. 
Motion, including the translational and rotational velocities of the \mbox{RGBlimp-Q}, is referenced to a body-fixed frame $O_b\text{-}x_by_bz_b$, with its origin $O_b$ defined at CB. 
The velocity reference frame $O_b\text{-}x_vy_vz_v$ is established to model aerodynamics, with its $O_b\text{-}x_v$ axis aligned with the direction of the translational velocity of the robot. 
The position vector $\boldsymbol{p}=[x,y,z]^\mathrm{T}$ describes the displacement of the origin of the inertial frame $O$ with respect to the origin of the body-fixed frame $O_b$. 
The Euler angles, including the roll angle $\phi\!\in\!\left(-\tfrac{\pi}{2}, \tfrac{\pi}{2}\right)$, pitch angle $\theta\!\in\!\left(-\tfrac{\pi}{2}, \tfrac{\pi}{2}\right)$, and yaw angle $\psi\!\in\!\left(-\pi, \pi\right)$, represent the orientation, denoted as $\boldsymbol{\eta}=[\phi,\theta,\psi]^\mathrm{T}$. 
Let $\boldsymbol{v}=[\dot{x},\dot{y},\dot{z}]^\mathrm{T}$ and $\boldsymbol{\omega}=[p,q,r]^\mathrm{T}$ represent the translational velocity expressed in the inertial frame and rotational velocity expressed in the body-fixed frame, respectively. 
As commonly used in the literature \cite{FlightDynamic}, rotation from the body-fixed frame to the inertial frame follows the roll-pitch-yaw (RPY) sequence, i.e., 
\begin{equation}
    \boldsymbol{R}=\boldsymbol{R}_z(\psi)\boldsymbol{R}_y(\theta)\boldsymbol{R}_x(\phi). \label{eq.4}
\end{equation}

The relationship between the time derivative of any given vector $\boldsymbol{c}$ expressed in the inertial frame $\dot{\boldsymbol{c}}^I$ and that expressed in the body-fixed frame $\dot{\boldsymbol{c}}^B$ follows $\footnote{Here $\dot{\boldsymbol{R}}=\boldsymbol{R} \boldsymbol{\omega}^\times$ is used for derivation, where $\boldsymbol{\omega}^\times$ is the skew-symmetric matrix corresponding to $\boldsymbol{\omega}$. }$
\begin{equation}
    \dot{\boldsymbol{c}}^I = \boldsymbol{R}\left(\dot{\boldsymbol{c}}^B + \boldsymbol{\omega}^B \times \boldsymbol{c}^B\right).
	\label{eq.5}
\end{equation}

To clarify the derivation, the motion dynamics model of the robot expresses all translational quantities in the inertial frame and all rotational quantities in the body-fixed frame.  
Where no ambiguity arises, the superscripts ``$I$\," and ``$B$\," are omitted, e.g., $\boldsymbol{v}\!=\!\boldsymbol{v}^I$ and $\boldsymbol{\omega}\!=\!\boldsymbol{\omega}^B$.

Then the kinematics of the \mbox{RGBlimp-Q} are described by
\begin{equation}
    \dot{\boldsymbol{p}}=\boldsymbol{v},\qquad \dot{\boldsymbol{\eta}}=\boldsymbol{P} \boldsymbol{\omega}. \label{eq.6}
\end{equation}
Here, the changing rate of the Euler angles $\dot{\boldsymbol{\eta}}$ and the rotational velocity $\boldsymbol{\omega}$ are related through the transformation matrix based on the RPY sequence, i.e., 
\begin{equation}
    \boldsymbol{P}=\left[\begin{array}{ccc}
        1 & \sin \phi \tan \theta & \cos \phi \tan \theta \\
        0 & \cos \phi & -\sin \phi \\
        0 & \sin \phi \sec \theta & \cos \phi \sec \theta
\end{array}\right].
	\label{eq.7}
\end{equation}

Based on Newton's second law of motion, the translational motion expressed in the inertial frame is described as
\begin{equation}
    \boldsymbol{F}_\mathrm{ext} = \hspace{0.5mm}m_0\left( \dot{\boldsymbol{v}} + \boldsymbol{R}\boldsymbol{a}^B \right), \label{eq.8}
\end{equation}
where $\boldsymbol{a}^B$ stands for the sum of Euler and centripetal acceleration expressed in the body-fixed frame, which is exerted by the point mass $m_0$ due to the displacement $\boldsymbol{r}_0$, defined as 
\begin{equation}
    \boldsymbol{a}^B= \dot{\boldsymbol{\omega}}\hspace{-0.6mm}\times\hspace{-0.6mm}\boldsymbol{r}_0+\boldsymbol{\omega}\hspace{-0.6mm}\times\hspace{-0.6mm}(\boldsymbol{\omega}\hspace{-0.6mm}\times\hspace{-0.6mm}\boldsymbol{r}_0) .
	\label{eq.9}
\end{equation}
The term $\boldsymbol{F}_\mathrm{ext}$ denotes the total external force acting on the stationary mass $m_0$ of the \mbox{RGBlimp-Q}, expressed in the inertial frame, calculated as
\begin{equation}
    \boldsymbol{F}_\mathrm{ext} = F\boldsymbol{R}\,\hat{\boldsymbol{i}} + (m_0g\!-\!F_b)\hat{\boldsymbol{k}} + \boldsymbol{R}\boldsymbol{F}_\mathrm{\!aero}^B+\boldsymbol{f}_a.
	\label{eq.10}
\end{equation}
Here, $F$ represents the forward thrust force from the propeller, aligned with the $x_{b'}$ axis as shown in Fig.\!~\ref{fig.4}. 
The vectors \mbox{$\hat{\boldsymbol{i}}\!=\![1,0,0]^\mathrm{T}$} and $\hat{\boldsymbol{k}}\!=\![0,0,1]^\mathrm{T}$ are unit vectors along the $O_b\text{-}x_b$ and $O\text{-}z$ axes, respectively. 
The scalar $g$ is the gravitational acceleration. 
The scalar $F_b\!=\!\rho gV_\text{He}$ represents the buoyant lift, where $\rho$ is the density of air and $V_\text{He}$ is the volume of helium gas.
The vector $\boldsymbol{F}_\text{\!aero}^{B}$ denotes the aerodynamic forces acting on the robot expressed in the body-fixed frame. 

The term $\boldsymbol{f}_a$ represents the total equivalent force acting on the stationary mass expressed in the inertial frame, exerted by the internal movable mass, which includes gravity and equivalent force, i.e., 
\begin{equation}
    \boldsymbol{f}_a = m_ag\hat{\boldsymbol{k}} - m_a\dot{\boldsymbol{v}}_a. \label{eq.11}
\end{equation}
Here, $\dot{\boldsymbol{v}}_a$ denotes the time derivative of the velocity of the internal movable mass $m_a$, expressed in the inertial frame, calculated as 
\begin{equation}
    \begin{aligned}
       \dot{\boldsymbol{v}}_a =&\ 
       \frac{\mathrm{d} \boldsymbol{v}_a}{\mathrm{d} t}=
       \frac{\mathrm{d}}{\mathrm{d} t}\boldsymbol{R}\left(\boldsymbol{v}^B + \boldsymbol{\nu}_a + \boldsymbol{\omega}\hspace{-0.8mm}\times\hspace{-0.6mm}\boldsymbol{r}_a\right) \\[1mm]
       =&\ \dot{\boldsymbol{v}} + \boldsymbol{R}\left(\dot{\boldsymbol{\nu}}_a+2\boldsymbol{\omega}\hspace{-0.6mm}\times\hspace{-0.6mm}\boldsymbol{\nu}_a +  \dot{\boldsymbol{\omega}}\hspace{-0.6mm}\times\hspace{-0.6mm}\boldsymbol{r}_a + \boldsymbol{\omega} \hspace{-0.6mm}\times\hspace{-0.6mm} \left( \boldsymbol{\omega}\hspace{-0.6mm}\times\hspace{-0.6mm}\boldsymbol{r}_a\right) \right), 
    \end{aligned} \label{eq.12}
\end{equation}
where $\boldsymbol{\nu}_a=\dot{\boldsymbol{r}}_a$ represents the time derivative of the controllable displacement $\boldsymbol{r}_a$ of the moving mass. 

Based on Euler's second axiom of motion, the rotational motion expressed in the inertial frame is described as
\begin{equation}
    \boldsymbol{T} = \boldsymbol{J}\dot{\boldsymbol{\omega}} + \boldsymbol{\omega}\hspace{-0.8mm}\times\hspace{-0.8mm}\left( \boldsymbol{J}\boldsymbol{\omega} \right) + m_0\boldsymbol{r}_0\hspace{-0.8mm}\times\hspace{-0.8mm}\boldsymbol{R}^\mathrm{T}\dot{\boldsymbol{v}}. \label{eq.13}
\end{equation}
Here, the matrix $\boldsymbol{J}\!\in\!\mathbb{R}^{3\!\times\!3}$ represents the inertia matrix of the stationary mass about the origin of the body-fixed frame. 
The total external moment $\boldsymbol{T}$ is calculated as
\begin{equation}
    \boldsymbol{T} = Fh\hat{\boldsymbol{j}} + \boldsymbol{r}_0\hspace{-0.6mm}\times\hspace{-0.6mm} m_0g\boldsymbol{R}^\mathrm{T}\hat{\boldsymbol{k}} + \boldsymbol{T}_\text{aero} + \boldsymbol{r}_a\hspace{-0.6mm}\times\hspace{-0.6mm}\boldsymbol{R}^\mathrm{T}\boldsymbol{f}_a. \label{eq.14}
\end{equation}
Here, the scalar $h$ denotes the distance between the forward thrust $F$ and the $O_b\text{-}x_b$ axis, as depicted in Fig.\!~\ref{fig.4}. 
The vector $\hat{\boldsymbol{j}}=[0,1,0]^\mathrm{T}$ represents the unit vector along the $O_b\text{-}y_b$ axis. 
The vector $\boldsymbol{T}_\text{aero}$ denotes the aerodynamic moment acting on the robot, expressed in the body-fixed frame. 

Let vector $\boldsymbol{F}_{a}\!=\!m_a\dot{\boldsymbol{\nu}}_a\!=\!m_a\ddot{\boldsymbol{r}}_a$ represent the control input force exerted on the internal moving mass $m_a$ in the body-fixed frame. 
Combining Eqs.\!~\eqref{eq.9}-\eqref{eq.14}, the motion dynamics of the \mbox{RGBlimp-Q} are expressed in terms of the control inputs including the forward thrust $F$ and the moving mass actuation force $\boldsymbol{F}_a$ as follows: 
\begin{equation}
\begin{aligned}
    \boldsymbol{E}
    \begin{bmatrix}
      \dot{\boldsymbol{v}}\\
      \dot{\boldsymbol{\omega}}\\
      \dot{\boldsymbol{\nu}}_{\!a}
    \end{bmatrix}
    &= 
    \begin{bmatrix}
      \tilde{\boldsymbol{F}}\\
      \tilde{\boldsymbol{T}}\\
      \boldsymbol{0}_{3\!\times\!1}
    \end{bmatrix} + \boldsymbol{N}\!
    \begin{bmatrix}
      F\\
      \boldsymbol{F}_a
    \end{bmatrix}, \label{eq.15}
\end{aligned}
\end{equation}
where
{\allowdisplaybreaks[4] 
\begin{IEEEeqnarray}{rCl}
    \tilde{\boldsymbol{F}}
    &=&
    \boldsymbol{R}(\boldsymbol{\omega}\!\times\!\boldsymbol{l}_g)\!\times\!\boldsymbol{\omega}\hspace{0.8mm}+((m_0\!+\!m_a)g\!-\!F_b)\hat{\boldsymbol{k}}\hspace{0.8mm}+\nonumber\\[-1.4mm]
    &&\boldsymbol{R}\boldsymbol{F}_\mathrm{aero} + 2m_a\boldsymbol{\nu}_a\!\times\!\boldsymbol{\omega}, \label{eq.16}\\
    \tilde{\boldsymbol{T}}\hspace{0.2mm}
    &=&
    \left(\boldsymbol{J}\!-\!m_a (\boldsymbol{r}_a^\times)^2\right)\boldsymbol{\omega}\!\times\!\boldsymbol{\omega} + \boldsymbol{l}_g\!\times\!g\boldsymbol{R}^\mathrm{T}\hat{\boldsymbol{k}} + \nonumber\\[-1.2mm]
    &&\boldsymbol{T}_\mathrm{aero} + 2m_a\boldsymbol{r}_a\!\times\!(\boldsymbol{\nu}_a\!\times\!\boldsymbol{\omega}), \label{eq.17}\\
    \boldsymbol{E}\hspace{0.2mm}&=&
    \begin{bmatrix}
      (m_0\!+\!m_a)\boldsymbol{I}&  -\boldsymbol{R}\,\boldsymbol{l}_g^\times & \boldsymbol{R}\\[+0.5mm]
      \boldsymbol{l}_g^\times\boldsymbol{R}^\mathrm{T}&  \boldsymbol{J}\!-\!m_a (\boldsymbol{r}_a^\times)^2&\, \boldsymbol{r}_a^\times\\[+0.5mm]
      \boldsymbol{0}_{3}&  \boldsymbol{0}_{3}& \boldsymbol{I}
    \end{bmatrix}, \label{eq.18}\\[1mm]
    \boldsymbol{N}\hspace{-0.15mm}&=&
    \begin{bmatrix}
     \ \boldsymbol{R}\hat{\boldsymbol{i}} & h\hat{\boldsymbol{j}} & \boldsymbol{0}_{1\!\times\!3}\ \\[+0.5mm]
     \multicolumn{2}{c}{\ \boldsymbol{0}_{3\!\times\!2}} & \tfrac{1}{m_a}\boldsymbol{I}
    \end{bmatrix}^{\mathrm{T}}. \label{eq.19}
\end{IEEEeqnarray}}
Here, $\boldsymbol{l}_g = m_0\boldsymbol{r}_0\!+\!m_a\boldsymbol{r}_a=(m_0\!+\!m_a)\boldsymbol{r}\!_g$ where $\boldsymbol{r}\!_g$ denotes the overall CG of the system of two point masses. 
$\boldsymbol{0}$ represents the zero vector or matrix. 
The matrix $\boldsymbol{I}\!\in\!\mathbb{R}^{\!3\!\times\!3}$ represents the identity matrix. 

Additionally, the aerodynamics terms in Eqs.\!~\eqref{eq.16} and \eqref{eq.17} need to be considered. 
Without loss of generality, we assume a windless environment, focusing on motion-induced aerodynamics. 
The transformation from the velocity reference frame $O_b\text{-}x_vy_vz_v$ to the body-fixed frame $O_b\text{-}x_by_bz_b$ is expressed as
\begin{equation}
    \boldsymbol{R}_v^b=\left[\begin{array}{ccc}
        \cos\alpha \cos\beta & -\cos\alpha \sin\beta & -\sin\alpha \\
        \sin\beta & \cos\beta & 0 \\
        \sin\alpha\cos\beta & -\sin\alpha \sin\beta & \cos\alpha
\end{array}\right].
	\label{eq.20}
\end{equation}
Here, the scalar $\alpha\!=\!\arctan(w/u)$ represents the angle of attack, and the scalar $\beta\!=\!\arcsin(v/V)$ denotes the sideslip angle, where the velocity magnitude $V\!=\!\left \| \boldsymbol{v} \right \|$ and the velocity expressed in the body-fixed frame $\boldsymbol{v}^B\!=\!\boldsymbol{R}^\mathrm{T}\boldsymbol{v}\!=\![u,v,w]^\mathrm{T}$. 

As illustrated in Fig.\!~\ref{fig.4}, the aerodynamic forces, including the drag force $D$, the side force $S$, and the lift force $L$, are defined in the velocity reference frame. 
Similarly, the aerodynamic moments comprise the roll moment $M_x$, the pitch moment $M_y$, and the yaw moment $M_z$. 
The aerodynamic forces and moments are described as
\begin{equation}
    \displaystyle
    \boldsymbol{F}_\text{aero}\!=\!\boldsymbol{R}_v^b[-D, S, -L]^\mathrm{T}\!\!\!,\quad
    \boldsymbol{T}_\text{aero}\!=\!\boldsymbol{R}_v^b[M_x, M_y, M_z]^\mathrm{T}\!\!\!,\label{eq.21}
\end{equation}
which are modeled as \scalebox{0.86}{$[D,\,S,\,L]^\mathrm{T}=\,\tfrac{1}{2}\rho V^2\!A[C_D, C_S, C_L]^\mathrm{T}$} and \scalebox{0.85}{$[M_{\!x},M_{\!y},M_{\!z}]^\mathrm{T}\!\!=\!\tfrac{1}{2}\rho V^2\!\!A[C_{\!M_{\!x}}, C_{\!M_{\!y}}, C_{\!M_{\!z}}]^\mathrm{T}\!\!+\![K_{\!x},K_{\!y},K_{\!z}]\,\boldsymbol{\omega}$}, where\! $A$ is the aerodynamic reference area. 
The terms with ``$C$'' and ``$K$'' denote the constant aerodynamic and damping coefficients, respectively, as experimentally identified in Section~\ref{Sec.sysID}. 

\subsection{Moving Mass Actuation via Continuum Mechanism}
\label{Sec.ContinnumModel}
As an essential design feature, the continuum mechanism implements moving mass actuation in the \mbox{RGBlimp-Q}. 
This section introduces the details and kinematics of the continuum arm without twist, based on the assumption of constant curvature (CC). 
Furthermore, we simplify the continuum arm with a lightweight and fixed-length elastic backbone (the central axis), thus a point mass $m_a$ on the end of the arm is considered, which is moving on a manifold with 2 DoF. 
\begin{figure}[thpb]
      \centering
      \includegraphics[width=88mm]{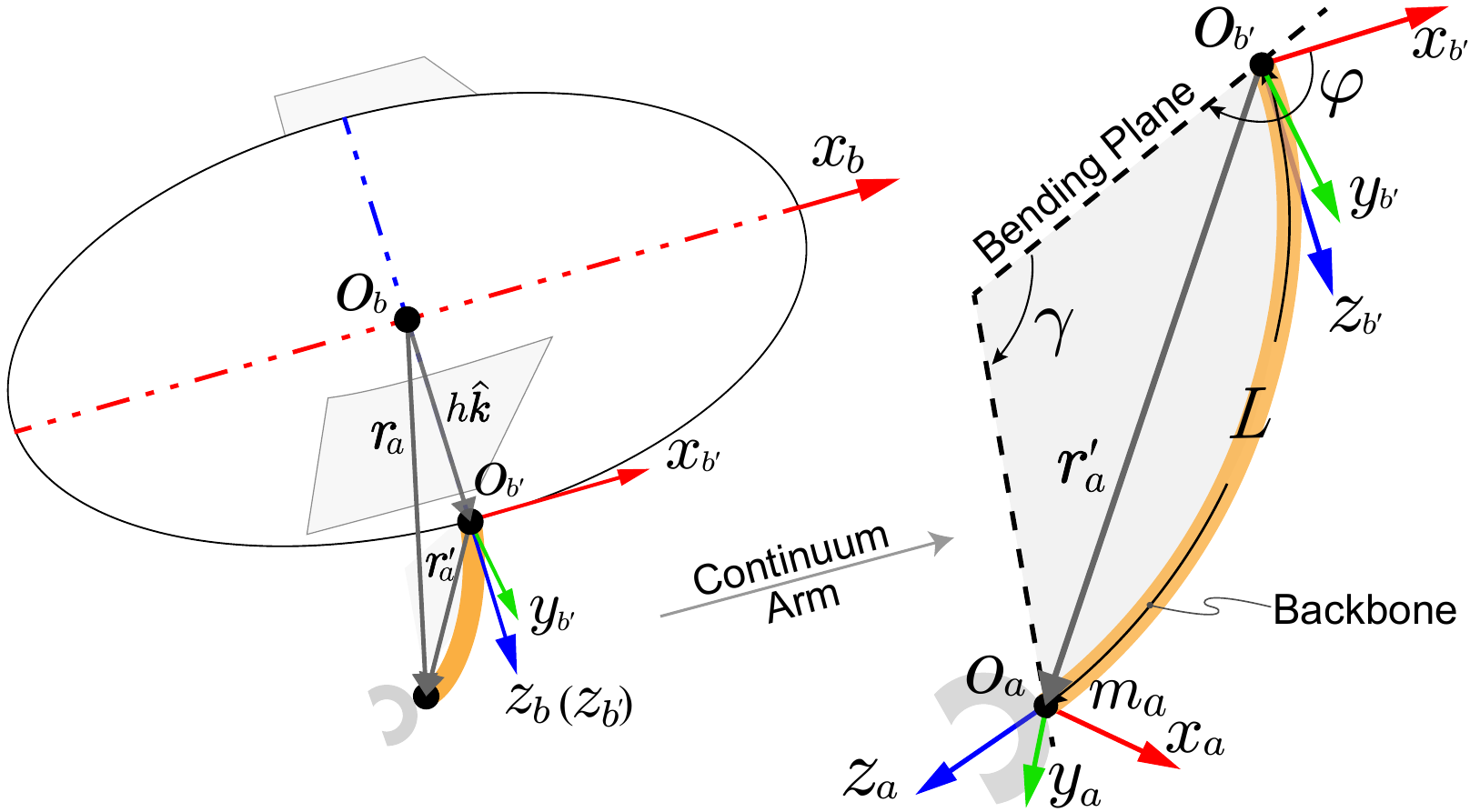}
      \caption{The geometric illustration of a continuum arm's spatial motion shows the arm bending in a plane, defined by the curvature angle $\gamma$, the bending direction $\varphi$, and reference frames: the base frame $O_{b'}\text{-}x_{b'}y_{b'}z_{b'}$ and the end-effector frame $O_a\text{-}x_ay_az_a$.} 
      \label{fig.5}
\end{figure}

Figure~\ref{fig.5} illustrates the CC model, where the continuum arm is approximated as a continuous segment with a curvature constant in space but varying in time. 
The base frame of the continuum arm $O_{b'}\text{-}x_{b'}y_{b'}z_{b'}$ has the same orientation as the body-fixed frame $O_b\text{-}x_by_bz_b$, but translated along $O_b\text{-}z_b$ axis with a distance $h$, as shown in Fig.\!~\ref{fig.5}. 
The end-effector frame of the continuum arm $O_a\text{-}x_ay_az_a$ is established at the end of the continuum arm. 
Under the CC assumption, the kinematic homogeneous transformation matrix $\boldsymbol{C}_{\!b'}^t\in\mathrm{SE}(3)$ maps arm $O_{b'}\text{-}x_{b'}y_{b'}z_{b'}$ into $O_a\text{-}x_ay_az_a$, given by 
\begin{equation}
    \boldsymbol{C}_{\!b}^t(\gamma,\varphi) = \boldsymbol{R}_z(\varphi)\!
       \begin{bmatrix}
         \boldsymbol{R}_x(\gamma)&\tfrac{L}{\gamma}\!
          \begin{bmatrix}
              \cos\gamma-\!1\\
              \sin\gamma\\
              0
          \end{bmatrix}\\[+1mm]
         0                       &  1
       \end{bmatrix}\!
       \boldsymbol{R}_z(-\varphi).
	\label{eq.22}
\end{equation}
Here, $L$ is the constant length of the continuum arm backbone.  
The angle $\varphi\!\in\![0,2\pi)$ and $\gamma\!\in\![0,\pi/2]$ stand for the direction of bending and the angle of curvature, respectively, as shown in Fig.\!~\ref{fig.5}. 
The matrices $\boldsymbol{R}_x, \boldsymbol{R}_z\!\in\!{\mathrm{SO}}(3)$ represent the rotation matrices about the $x_{b'}$ and $z_{b'}$ axes, respectively. 

We examine the translation vector from $O_{b'}\text{-}x_{b'}y_{b'}z_{b'}$ to $O_a\text{-}x_ay_az_a$, denoted as $\boldsymbol{r}'_a$. 
Derived from the transformation matrix \eqref{eq.22}, the expression for $\boldsymbol{r}'_a$ is given by
\begin{equation}
    \boldsymbol{r}'_a(\varphi,\gamma) = \frac{L}{\gamma}
    \begin{bmatrix}
      \ \cos\varphi(1-\cos\gamma)\ \\[+0.5mm]
      \ \sin\varphi(1-\cos\gamma)\ \\[+0.5mm]
      \sin\gamma
    \end{bmatrix}.
	\label{eq.23}
\end{equation}
As shown in Fig.\!~\ref{fig.4}, the translation vector $\boldsymbol{r}'_a$ plays a crucial role in the moving mass control, which determines the controllable displacement $\boldsymbol{r}_a$, i.e., $\boldsymbol{r}_a = [0,0,h]^\mathrm{T} + \boldsymbol{r}'_a$. 

The configuration $[\gamma,\varphi]^\mathrm{T}$ represents a standard parametrization commonly used in continuum robot studies \cite{ContinuumPCC,ContinuumSurvey3}. 
To address the inherent limitations of this approach, Santina and Rus proposed a novel $q$-parametrization for continuum arms, which are actuated by four evenly distributed cables and have no twist. 
This parametrization is defined as linear combinations of the four cable lengths \cite{QParam}. 
For a lightweight-oriented design that also facilitates spatial motion, we consider a configuration with three cables---the minimum required for a single-segment cable-driven spatial continuum. 
Furthermore, we derive the \mbox{$q$-parametrization} for this three-cable configuration, establishing its kinematic mapping relationship. 
\begin{figure*}[thpb]
      \centering
      \includegraphics[scale=1]{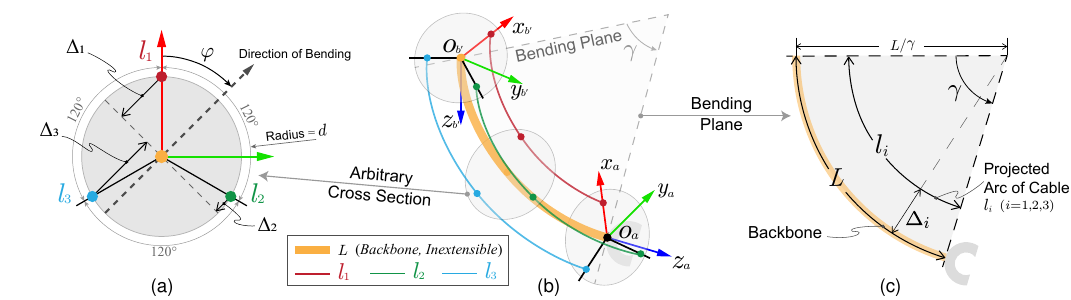}
      \caption{Illustration of the $q$-parametrization of the three-cable-based continuum arm. 
               The parameters $q=[\delta_x,\delta_y]^\mathrm{T}\in\mathbb{R}^2$ are defined as a linear combination of the lengths of the three evenly distributed cables $(l_1,l_2,l_3)$. 
               (a) depicts the geometric relation in arbitrary cross section, where the cables are connected without twist and are equidistant, each at a distance $d$ from the center. 
               (b) illustrates the entire continuum body deformed by the three cables. 
               (c) shows the bending plane from a perpendicular view, visually illustrating the projected arc of any cable $l_i, (i\!=\!1,2,3)$ at corresponding projection distance $\Delta_i$, as (a). 
               } 
      \label{fig.6}
\end{figure*}

Figure~\ref{fig.6} illustrates the $q$-parametrization of the three-cable-driven continuum arm, 
featuring three evenly distributed cables $l_1$, $l_2$, and $l_3$ aligned parallel to the backbone in the bending plane of the continuum arm. 
These cables share the same curvature as the backbone, maintaining a constant distance $d$ from the center in an arbitrary cross section of the continuum arm, as shown in the right plot of Fig.\!~\ref{fig.6}. 
Trigonometric considerations allow us to derive the projection distances on the bending plane $\Delta_i\ (i\!=\!1,2,3)$ of the constant distance $d$ for each cable, i.e., 
\begin{equation}
    \displaystyle
    \Delta_1\!=\!+d \cos\varphi,\quad  \Delta_{2,3}\!=\!-d \sin(\tfrac{\pi}{6}\!\mp\!\varphi).\label{eq.24}
\end{equation}
Further, the length of cable $l_i, (i\!=\!1,2,3)$ with the projection distance $\Delta_i$, as shown in Fig.\!~\ref{fig.6}, is calculated by 
\begin{equation}
    \hspace{0.5cm}l_i = \gamma(\tfrac{L}{\gamma}-\Delta_i) = L - \gamma\Delta_i,\hspace{0.3cm}(i=1,2,3). \label{eq.25}
\end{equation}

Combining Eqs.\!~\eqref{eq.24} and \eqref{eq.25} yields the lengths of the three cables as functions of $\gamma$ and $\varphi$, i.e.,
\begin{align}
    \displaystyle
    l_1(\gamma,\varphi)&= L - d\gamma\cos\varphi,\nonumber\\
    l_2(\gamma,\varphi)&= L + d\gamma\sin(\tfrac{\pi}{6}\!-\!\varphi),\label{eq.26}\\
    l_3(\gamma,\varphi)&= L + d\gamma\sin(\tfrac{\pi}{6}\!+\!\varphi).\nonumber
\end{align}

\noindent Furthermore, we define the new state configuration as linear combinations of the three cable lengths, as expressed through the $q$-parametrization, i.e.,
\begin{align}
    \displaystyle
    \delta_x  &=\tfrac{1}{3}(l_2\!+\!l_3-2l_1)=d\gamma\cos\phi,\nonumber\\
    \delta_y  &=\tfrac{\sqrt{3}}{3}(l_3\!-\!l_2)=d\gamma\sin\phi.\label{eq.27}\\
    \delta\,\,&=\sqrt{\delta_x^2\!+\!\delta_y^2}=d\gamma\nonumber
\end{align}
Let $q = [\delta_x,\delta_y]^\mathrm{T}\in\mathbb{R}^2$ represent the continuum configuration in $q$-parametrization, describing the bending degree in the $x_{b'}$ and $y_{b'}$ directions, respectively. 
The mapping from the $q$-parametrization to the angle parametrization is described as 
\begin{equation}
    \varphi = \arccos(\frac{\delta_x}{\delta})= \arcsin(\frac{\delta_y}{\delta}),\qquad  \gamma = \frac{\delta}{d}. \label{eq.28}
\end{equation}

Let the velocity in $q$-parametrization serve as the control input, denoted as $\dot{\delta}_x$ and $\dot{\delta}_y$. 
Such input is well-suited for a cable-driven continuum arm, discussed in detail in Section~\ref{Sec.CableDriven}. 
Combining Eqs.\!~\eqref{eq.23} and \eqref{eq.28}, the translation vector $\boldsymbol{r}'_a$ in the $q$-parametrization is given by
\begin{equation}
    \boldsymbol{r}'_a(\delta_x,\delta_y) = \frac{Ld}{\delta^2}
    \begin{bmatrix}
      \ \delta_x(1-\cos\frac{\delta}{d})\ \\[+1mm]
      \ \delta_y(1-\cos\frac{\delta}{d})\ \\[+1mm]
        \delta\sin\frac{\delta}{d}
    \end{bmatrix}.
	\label{eq.29}
\end{equation}

Therefore, the time derivative of the displacement vector $\boldsymbol{r}_a\!=\![0,0,h]^\mathrm{T}\!+\!\boldsymbol{r}'_a$ is determined as 
\begin{equation}
    \dot{\boldsymbol{r}}_a =  \frac{\partial \boldsymbol{r}'_a}{\partial (\delta_x, \delta_y)} 
    \begin{bmatrix}
        \dot{\delta}_x\\[+0.5mm]
        \dot{\delta}_y
    \end{bmatrix}, 
	\label{eq.30}
\end{equation}
where ${\partial \boldsymbol{r}'_a}/{\partial (\delta_x, \delta_y)}\!\in\!\mathbb{R}^{3\!\times\!2}$ denotes the Jacobian matrix of the RHS of Eq.\!~\eqref{eq.30}.

\subsection{Full Dynamic Model of RGBlimp-Q}
\label{Sec.full}
Combining the motion dynamics \eqref{eq.16} and the kinematics of the continuum arm \eqref{eq.30}, we derive the full dynamic model of the \mbox{RGBlimp-Q}, expressed in terms of the system states $\boldsymbol{p}, \boldsymbol{\eta}, \boldsymbol{v}, \boldsymbol{\omega}, \boldsymbol{r}_a$ and the control inputs $F, \dot{\delta}_x, \dot{\delta}_y$, i.e.,
\begin{align}
    \begin{bmatrix}
      \dot{\boldsymbol{p}}\\[+0.35mm]
      \dot{\boldsymbol{\eta}}\\[+0.35mm]
      \,\dot{\boldsymbol{r}}_{\!a}\,\\[+0.35mm]
      \dot{\boldsymbol{v}}\\[+0.35mm]
      \dot{\boldsymbol{\omega}}
    \end{bmatrix}\hspace{1mm}
    =\ 
    \begin{bmatrix}
      \boldsymbol{v}\\
      \boldsymbol{P} \boldsymbol{\omega}\\
      \boldsymbol{0}_{3\!\times\!1}\\[+1mm]
      \boldsymbol{A}\!
    \begin{bmatrix}
      \,\tilde{\boldsymbol{f}}\,\,\\
      \,\tilde{\boldsymbol{t}}\,\,
    \end{bmatrix}
    \end{bmatrix}+
    \boldsymbol{B}\!
    \begin{bmatrix}
      \,F\,\\
      \,\dot{\delta}_x\,\\
      \,\dot{\delta}_y\,
    \end{bmatrix},\label{eq.31}
\end{align}
where
\begin{align}
    \hspace{1mm}\tilde{\boldsymbol{f}}\hspace{0.5mm} 
    &= \boldsymbol{R}(\boldsymbol{\omega}\!\times\!\boldsymbol{l}_g)\!\times\!\boldsymbol{\omega}+((m_0\!+\!m_a)g\!-\!F_b)\hat{\boldsymbol{k}} + \boldsymbol{R}\boldsymbol{F}_\mathrm{aero}, \label{eq.32}\\[+1.5mm]
    \hspace{1mm}\tilde{\boldsymbol{t}}\hspace{0.8mm} 
    &= \left(\boldsymbol{J}\!-\!m_a (\boldsymbol{r}_a^\times)^2\right)\!\boldsymbol{\omega}\!\times\!\boldsymbol{\omega} + \boldsymbol{l}_g\!\times\!g\boldsymbol{R}^\mathrm{T}\hat{\boldsymbol{k}} + \boldsymbol{T}_\mathrm{aero}, \label{eq.33}
\end{align}
\begin{equation}
    \boldsymbol{A}=\!
    \begin{bmatrix}
      (m_0\!+\!m_a)\boldsymbol{I}&  -\boldsymbol{R}\,\boldsymbol{l}_g^\times\\[+0.5mm]
      \boldsymbol{l}_g^\times\boldsymbol{R}^\mathrm{T}&  \boldsymbol{J}\!-\!m_a (\boldsymbol{r}_a^\times)^2
    \end{bmatrix}^{\!-\!1}\!\!\!,\hspace{3em} \label{eq.34}
\end{equation}
\begin{equation}
    \boldsymbol{B}=\!
    \begin{bmatrix}
        \multicolumn{2}{c}{\boldsymbol{0}_{6\!\times\!4}}\\[+0.5mm]
        \quad\quad\boldsymbol{0}_{3\!\times\!1}  & \boldsymbol{I}\  \\[+0.5mm]
        \multicolumn{2}{c}{\boldsymbol{A}
        \begin{bmatrix}
            \hat{\boldsymbol{i}} & -2\boldsymbol{\omega}^\times\\
            h\hat{\boldsymbol{j}} & -2\boldsymbol{r}_a^\times\boldsymbol{\omega}^\times
        \end{bmatrix}}
    \end{bmatrix}
    \begin{bmatrix}
        1 & \boldsymbol{0}_{1\!\times\!2} \\
        \boldsymbol{0}_{3\!\times\!1} &  \frac{\partial \boldsymbol{r}'_a}{\partial (\delta_x, \delta_y)} 
    \end{bmatrix}.\hspace{-0.78em}\label{eq.35}
\end{equation}
Here, the allocation matrix $\boldsymbol{A}$ is always invertible (see \mbox{Appendix~\ref{Appendix.A}} for proof), ensuring that $\boldsymbol{A}^{-1}$ is always well-defined. 
The inertial matrix $\boldsymbol{J}-m_a (\boldsymbol{r}_a^\times)^2$ in Eq.\!~\eqref{eq.34} represents the effective rotational inertia of the entire system, including both the stationary mass $m_0$ and the moving mass $m_a$, i.e., 
\begin{equation}
    \begin{aligned}
        \boldsymbol{J}_\mathrm{eff} &= \boldsymbol{J}-m_a (\boldsymbol{r}_a^\times)^2 \\
        &=\boldsymbol{J}+m_a(\|\boldsymbol{r}_a\|^2 \boldsymbol{I} - \boldsymbol{r}_a \boldsymbol{r}_a^\mathrm{T}).
    \end{aligned}\label{eq.36}
\end{equation}
The design of \mbox{RGBlimp-Q} ensures that $\boldsymbol{r}_a \neq \boldsymbol{0}$ and $\boldsymbol{J} \succ 0$, which guarantees that $\det(\boldsymbol{J}_\mathrm{eff}) > 0$.  
This ensures that the dynamics described by Eq.\!~\eqref{eq.31} remain well-posed (i.e., avoiding ill-conditioning), and that rotational sensitivity remains bounded. 
For example, the roll angular acceleration $\ddot{\phi}={\tau_x}/{J_{xx}}$ remains finite as long as $J_{xx}>0$, where $\tau_x$ denotes the torque around the $x$-axis. 

\section{Continuum Arm System}
\label{Sec.continuum}
\begin{figure}[htbp]
      \centering
      \includegraphics[width=88mm]{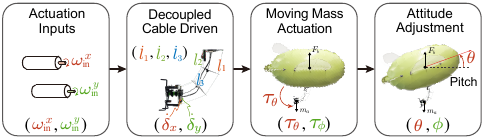}
      \caption{The schematic diagram illustrates the proposed decoupled mechanism, which enables decoupled cable-driven control of the continuum arm and facilitates moving mass actuation for attitude adjustment of \mbox{RGBlimp-Q}. 
       The decoupled mechanism enables the mechanism-level decoupled control of the continuum arm for 2-DoF spatial motion using two actuators. 
       Specifically, a compound gear train is designed to coordinate the adjustment of three cables under continuum constraints, thereby realizing the 2-DoF spatial motion required for moving mass actuation. This motion facilitates attitude adjustment of \mbox{RGBlimp-Q} in both pitch $\theta$ and roll $\phi$ directions.}
      \label{fig.7}
      \vspace{5pt}
\end{figure}

In this section, we present the design of a decoupled, cable-driven continuum arm with an end-effector claw, specifically developed to support the proposed moving-mass-based attitude adjustment of \mbox{RGBlimp-Q}, see Fig.\!~\ref{fig.7}. 

The mechanism enables decoupled control of the continuum arm to achieve 2-DoF spatial motion using only two actuators. 
This design eliminates the need for redundant actuators, thereby significantly reducing both system weight and energy consumption---critical considerations for payload-constrained robotic blimp platforms.

\subsection{Mechanically Decoupled Cable-Driven Design}
\label{Sec.CableDriven}
In our design, we consider the continuum arm without twist under the CC assumption, as discussed in Section~\ref{Sec.ContinnumModel} and illustrated in Fig.\!~\ref{fig.6}, where the cables are treated as the driving cables.
To achieve spatial motion of the continuum arm, a minimum of three cables is required.  
These cables are constrained by the backbone, resulting in a two-degree-of-freedom motion described in terms of either $(\gamma,\varphi)$ or $(\delta_x,\delta_y)$. 
In the literature \cite{ContinuumSurvey,ContinuumSurvey2,ContinuumSurvey3}, these cables are typically controlled by independent actuators (motors) within specific constraints. 
However, such redundancy in motors leads to increased inefficiency in load handling, a concern that is particularly pertinent in the design of miniature aircraft such as the \mbox{RGBlimp-Q}. 
To mitigate this inefficiency, we utilize a compound gear train mechanism to achieve decoupling for the cable-driven continuum arm. 
This approach ensures that the two-degree-of-freedom motion of the robot is independently controlled by two corresponding actuators. 

Let's reconsider the relationship between three cable lengths $(l_1,l_2,l_3)$ and the $q$-parametrization configuration $(\delta_x, \delta_y)$. 
Combining Eqs.\!~\eqref{eq.26} and \eqref{eq.27}, we derive the lengths of the three cables as functions of $\delta_x$ and $\delta_y$, i.e.,
\begin{align}
    \begin{bmatrix}
      \,l_1\,\\[+0.15mm]
      \,l_2\,\\[+0.15mm]
      \,l_3\,
    \end{bmatrix}\hspace{1mm}
    =\ 
    \begin{bmatrix}
      -1              & \ \ 0 \\[+0.2mm]
      \hspace{2.7mm}1 & - \tfrac{\sqrt{3}}{2}\,\,\,\\[+0.2mm]
      \hspace{2.7mm}1 & \hspace{2.7mm} \tfrac{\sqrt{3}}{2}\,\,\,
    \end{bmatrix}
    \begin{bmatrix}
      \delta_x\\[+0.15mm]
      \delta_y
    \end{bmatrix}.\label{eq.37}
\end{align}

These equations reveal that the lengths of the three cables are kinematically constrained by a linear relationship, i.e., $l_1+l_2+l_3=0$. 
Further, it is observed that the length $l_1$ is independent of $\delta_y$ but decreases with $\delta_x$. 
Meanwhile, $l_2$ and $l_3$ increase with $\tfrac{1}{2}\delta_x$. 
The length $l_2$ decreases with $\tfrac{\sqrt{3}}{2}\delta_y$, while $l_3$ increases with $\tfrac{\sqrt{3}}{2}\delta_y$. 

\begin{figure}[thpb]
      \centering
      \includegraphics[width=88mm]{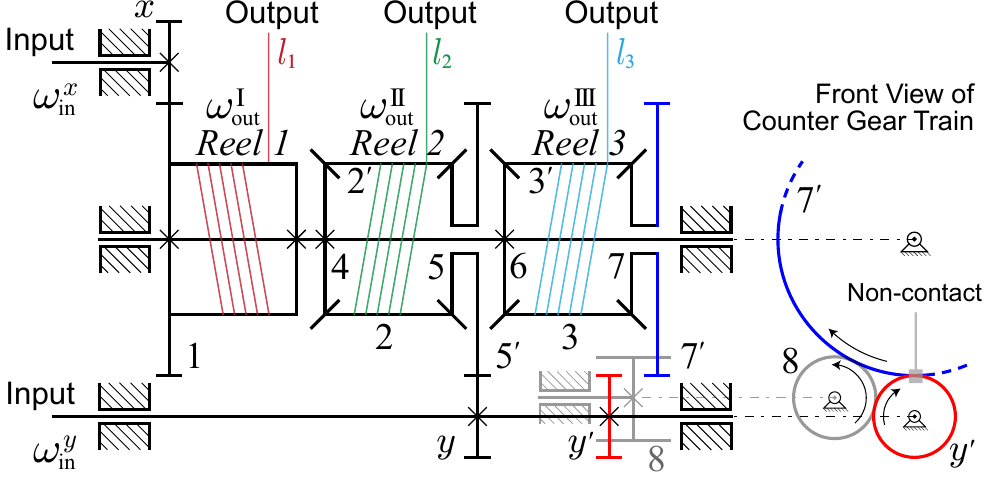}
      \caption{Representation of the proposed compound-gear-train-based decoupling mechanism. 
               The design ensures independent control for the 2 DoF of the continuum arm by two actuators accordingly. 
               The three cables are winded at each of three outputting reels, including differential-based reels $2$ and $3$, and the fixed-axis reel $1$ with an opposite winding direction. 
               The left part of the figure shows the main structure of the mechanism in the side view. 
               The right part depicted the front view of the counter gear, i.e., gear $8$, whose rotating axis is in a different plane.  
               Note that the gear $7'$ (blue) and $y'$ (red) are not engaged, as shown in the side view.}
      \label{fig.8}
\end{figure}

Based on Eq.\!~\eqref{eq.37}, we propose the mechanism of a compound gear train for decoupled cable driving, as shown in \mbox{Fig.\!~\ref{fig.8}}. 
The mechanism uses two actuators for decoupled inputs, yielding three rotary motions for winding the three cables under the constraint in Eq.\!~\eqref{eq.37} accordingly. 
The inputs and outputs of the mechanism are defined as the rotational speeds of the two actuators and the three reels, respectively, denoted as $(\omega_\text{in}^x,\omega_\text{in}^y)$ and $(\omega_\text{out}^\mathrm{I}, \omega_\text{out}^\mathrm{\ii}, \omega_\text{out}^\mathrm{\iii})$. 
Specifically, reel $1$ is directly affixed to a shaft, i.e., the fixed-axis reel, while reel $2$ and reel $3$ are made of differential gear trains, i.e., differential-based reels. 
The setup includes seven spur gears ($x$, $y$, $y'$\!, $1$, $5'$\!, $7'$\!, $8$) and eight bevel gears ($2$, $2'$\!, $3$, $3'$\!, $4$, $5$, $6$,~$7$). 
To determine the overall drive ratio $R$ of the compound gear train, we examine the rotational speeds $\omega$ and tooth numbers $n$ of each gear. 
Beginning with the fixed-axis gear train segment, we compute the rotational speed of reel $1$ as
\begin{equation}
    \displaystyle
    \omega_\text{out}^\mathrm{I} = R_x^1\omega_\text{in}^x= -\tfrac{n_x}{n_1}\omega_\text{in}^x. \label{eq.38}
\end{equation}
Here, $R_x^1$ represents the drive ratio from gear~$x$ to gear~$1$, with the negative sign indicating the opposite direction of rotation. 
In addition, gear~$4$ and gear~$6$ are fixed on the same shaft with gear~$1$, resulting in $\omega_4=\omega_6=\omega_1$. 
Similarly, we obtain 
\begin{equation}
    \displaystyle
    \omega_5 = \omega_{5'} = -\tfrac{n_y}{n_{5'}}\omega_\text{in}^y, \quad \omega_7=\omega_{7'} = \tfrac{n_8}{n_{7'}}\tfrac{n_{y'}}{n_8}\omega_\text{in}^y.\label{eq.39}
\end{equation}

The tooth numbers of bevel gears in the differential gear train are designed the same, i.e., $n_4\!=\!n_5\!=\!n_6\!=\!n_7$.
Thus, the rotational speeds of reels $2$ and $3$ are calculated as 
\begin{equation}
    \displaystyle
    \omega_\text{out}^\mathrm{\ii} = \tfrac{1}{2}(\omega_4+\omega_5),\quad \omega_\text{out}^\mathrm{\iii} = \tfrac{1}{2}(\omega_6+\omega_7). \label{eq.40}
\end{equation}

Combining Eqs.\!~\eqref{eq.38}, \eqref{eq.39} and \eqref{eq.40}, we obtain the rotational speeds of reels $2$ and $3$ as 
\begin{equation}
    \displaystyle
    \omega_\text{out}^\mathrm{\ii}=-\tfrac{1}{2}\tfrac{n_x}{n_1}\omega_\text{in}^x-\tfrac{1}{2}\tfrac{n_y}{n_{5'}}\omega_\text{in}^y,\hspace{0.5em}
    \omega_\text{out}^\mathrm{\iii}=-\tfrac{1}{2}\tfrac{n_x}{n_1}\omega_\text{in}^x+\tfrac{1}{2}\tfrac{n_{y'}}{n_{7'}}\omega_\text{in}^y. \nonumber
\end{equation}
Here, we selected two different sets of tooth numbers for spur gears, i.e., $n_x\!=\!n_y\!=\!n_{y'}\!=\!n_8$ and $n_1\!\!=\!n_{5'}\!\!=\!n_{7'}$. 
Combining Eq.\!~\eqref{eq.45} with $k\!=\!{n_x}/{n_1}\!=\!{n_y}/{n_{5'}}\!=\!{n_{y'}}\!/{n_{7'}}$, thus
\begin{equation}
    \omega_\text{out}^\mathrm{I}\!=\!-k \omega_\text{in}^x,\  
    \omega_\text{out}^\mathrm{\ii}\!=\!-\tfrac{1}{2}k(\omega_\text{in}^x\!+\!\omega_\text{in}^y),\ 
    \omega_\text{out}^\mathrm{\iii}\!=\!-\tfrac{1}{2}k(\omega_\text{in}^x\!-\!\omega_\text{in}^y). \nonumber
\end{equation}

Considering that cable $l_1$ is wound in the opposite direction to $l_2$ and $l_3$, and the radii of the reels are designed the same, denoted as $r_\text{reel}$,
the velocities of the cables are calculated as
\begin{align}
    \begin{bmatrix}
      \,\dot{l}_1\,\\[+0.35mm]
      \,\dot{l}_2\,\\[+0.35mm]
      \,\dot{l}_3\,
    \end{bmatrix}\hspace{1mm}
    =\ 
    \begin{bmatrix}
      -1              & \ \ 0 \\[+0.35mm]
      \hspace{2.7mm}1 & - \tfrac{\sqrt{3}}{2}\,\,\,\\[+0.35mm]
      \hspace{2.7mm}1 & \hspace{2.7mm} \tfrac{\sqrt{3}}{2}\,\,\,
    \end{bmatrix}
    \begin{bmatrix}
      k r_\text{reel}\omega_\text{in}^x\\[+0.5mm]
      \tfrac{\sqrt{3}}{3}k r_\text{reel}\omega_\text{in}^y
    \end{bmatrix}.\label{eq.41}
\end{align}

Comparing Eqs.\!~\eqref{eq.37} and \eqref{eq.41}, we find
\begin{equation}
    \displaystyle
    \dot{\delta}_x = k r_\text{reel} \omega_\text{in}^x,\quad \dot{\delta}_y = \tfrac{\sqrt{3}}{3}k r_\text{reel} \omega_\text{in}^y. \label{eq.42}
\end{equation}

By considering $(\omega_\text{in}^x, \omega_\text{in}^y)$ as inputs, the decoupled cable-driven mechanism facilitates independent control of the 2-DoF continuum arm using two actuators, thereby mitigating the complexity of controlling the three cables. 

\subsection{Moving-Mass-Based Attitude Control}
\label{Sec.Control}
The decoupled cable-driven mechanism enables independent control of the continuum-based moving mass in 2 DoF through inputs ($\omega_\text{in}^x,\omega_\text{in}^y$), thereby adjusting attitude in pitch ($\theta$) and roll ($\phi$). 
In this section, we propose an attitude controller for the \mbox{RGBlimp-Q} based on a classic PID feedback framework and demonstrate its stability. 
This controller is implemented to demonstrate the inherent advantages of the novel continuum-based moving mass actuation mechanism design in enabling effective attitude adjustment and efficient power utilization, as illustrated by experiments in Section~\ref{Sec.experiment}. 

\begin{figure}[thpb]
      \centering
      \includegraphics[scale=0.46]{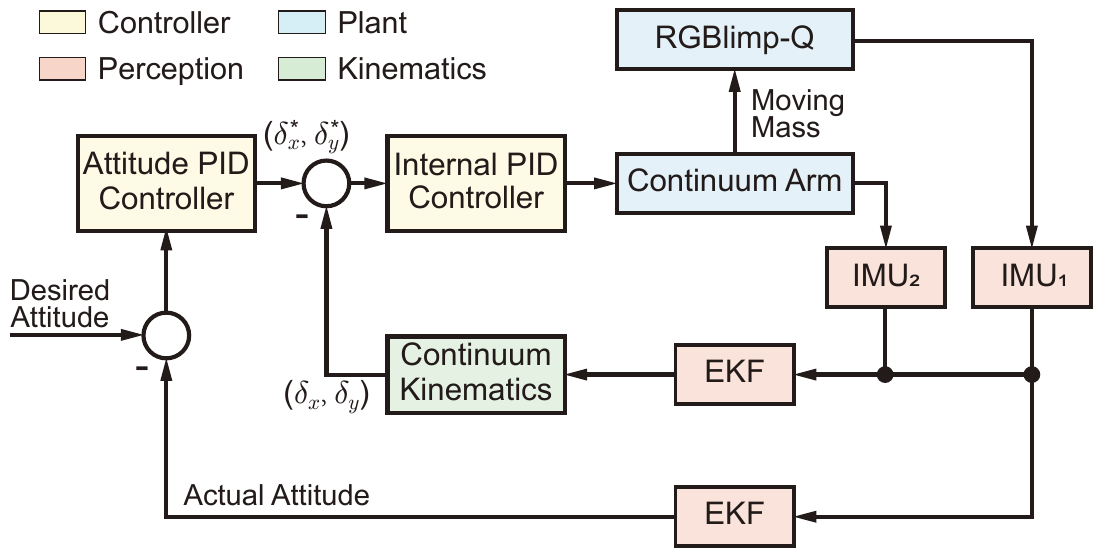}
      \caption{Control schematic of the dual-loop PID. 
               The internal controller regulates the continuum arm configuration in the $q$-parametrization, i.e., $(\delta_x,\delta_y)$. 
               The attitude controller adjusts the orientation of \mbox{RGBlimp-Q}, specifically the roll, pitch, and yaw angles, utilizing both the gravitational moment of the moving mass and the aerodynamic moments.}
      \label{fig.9}
\end{figure}

Two IMUs are employed as feedback sensors, mounted at both ends of the continuum arm, specifically the base frame $O_{b'}\text{-}x_{b'}y_{b'}z_{b'}$ and the end-effector frame $O_a\text{-}x_ay_az_a$ of the continuum arm. 
Leveraging extended Kalman filters (EKF) and the kinematics of the continuum arm Eqs.\!~\eqref{eq.28}-\eqref{eq.30}, these sensors provide crucial information regarding the robot's attitude and the arm configuration.  
Figure~\ref{fig.9} illustrates the schematic of the feedback controller. 

Overall, the control objectives encompass tracking the desired configuration of the continuum arm and maintaining the desired flight attitude. 
To achieve this, the controller is designed into an inner loop and an outer loop, aligning with these two objectives, respectively. 
The inner-loop controller governs the movement of the two motors controlling the continuum arm, thereby managing the tracking error in its configuration $(\delta_x,\delta_y)$.  
More importantly, the outer-loop controller adjusts the robot's attitude by utilizing the gravitational moment induced by the moving mass for the pitch angle $\theta$ and the roll angle $\phi$, as well as the aerodynamic yaw moment generated by airflow for the yaw angle $\psi$. 

For clarity, we analyze pitch attitude tracking stability under moving mass actuation with $\delta_x$ as active control input and $\delta_y$ fixed. 
The pitch dynamics of \mbox{RGBlimp-Q} are represented by
\begin{equation}
    J_{yy}^{\text{eff}}(\delta_x) \, \ddot{\theta} = \tau_\theta(\delta_x) - D_\theta \dot{\theta} + \Delta_\theta. \label{eq.43}
\end{equation}
Here, $\tau_\theta = m_a g r_{\!a\hspace{-0.3mm},x}$ is the gravitational moment generated by the continuum-based moving mass, 
$D_\theta\!>\!0$ represents the pitch damping coefficient, and $\Delta_\theta$ is a bounded disturbance with $|\Delta_\theta|\!\leq\!\rho_\theta$, where $\rho_\theta\!>\!0$ denotes the upper bound of the pitch disturbance. 
$J_{yy}^{\text{eff}}=\!J_{yy} + m_a ( r_{\!a\hspace{-0.3mm},x}^2\!+\!r_{\!a\hspace{-0.3mm},z}^2)$ denotes the corresponding diagonal component of the effective rotational inertia tensor $\boldsymbol{J}_\mathrm{eff}\!=\!\boldsymbol{J}+m_a(\|\boldsymbol{r}_a\|^2 \boldsymbol{I} - \boldsymbol{r}_a \boldsymbol{r}_a^\mathrm{T})$, where $r_{\!a\hspace{-0.3mm},x}\!=\!\frac{Ld}{\delta_x} (1\!-\!\cos\frac{\delta_x}{d})$ and $r_{\!a\hspace{-0.3mm},z}\!=\!\frac{Ld}{\delta_x} \sin\frac{\delta_x}{d}$ from the kinematics of the continuum arm. 
For simplicity, we assume $\delta_x \ll d$, thus 
\begin{equation}
    r_{a,x} \approx \frac{L \delta_x}{2d}, \quad r_{a,z} \approx h + L. \label{eq.44}
\end{equation}
The effective inertia is simplified as
\begin{equation}
    J_{yy}^{\text{eff}}(\delta_x) \approx J_{yy} + m_a \left( \frac{L^2 \delta_x^2}{4d^2} + \left(h + L\right)^2 \right), \label{eq.45}
\end{equation}
with $\dot{J}_{yy}^{\text{eff}} \approx \frac{m_a L^2}{2d^2} \delta_x \dot{\delta}_x$ and $\tau_\theta \approx -\frac{m_a g L}{2d} \delta_x$. 

Given the desired pitch angle $\theta^*$, the outer-loop pitch attitude control law is $\delta_x\!=\!-k_p e_\theta\!-\!k_d \dot{e}_\theta$, where $k_p,k_d\!>\!0$ are the control gains, $e_\theta\!=\!\theta\!-\!\theta^*$ represents the tracking error, and $\dot{e}_\theta\!=\!\dot{\theta}$. 
Considering time-varying inertia $J_{yy}^{\text{eff}}(\delta_x)$ and disturbances $\Delta_\theta$, we define the following quadratic function as the Lyapunov function for stability analysis, i.e.,
\begin{equation}
    V(\boldsymbol{x}) = \frac{1}{2} J_{yy}^{\text{eff}} \dot{e}_\theta^2 + \frac{1}{2} k_pe_\theta^2 + \frac{\lambda}{4}\delta_x^2, \label{eq.46}
\end{equation}
where $\frac{\lambda}{4}\delta_x^2$ represents control energy with tunable parameter $\lambda\!>\!0$. 
Combining both the pitch dynamics \eqref{eq.43} and the control law, the time derivative $\dot{V}$ is calculated and simplified as
\begin{align}
    \textstyle
    \dot{V} &\textstyle= -\left( D_{\theta} + \frac{m_a g L k_d}{2d} \right) \dot{e}_{\theta}^2 - \frac{\lambda_{k_p}}{2} e_{\theta}^2 - \frac{\lambda_{k_d}}{2} \dot{e}_{\theta}^2 + \nonumber\\
            &\textstyle\hspace{1em}\tfrac{1}{2}\!\left(\tfrac{m_a L^2}{d^2} \delta_x \dot{\delta}_x \dot{e}_\theta^2 - {\lambda k_p} e_\theta \dot{e}_\theta \right)+ \dot{e}_{\theta} \Delta_{\theta}.
    \label{eq.47}
\end{align}
Applying Young's inequality to the disturbance $\dot{e}_{\theta} \Delta_{\theta}$ yields
\begin{equation}
    \dot{e}_\theta \Delta_\theta \leq \frac{\epsilon}{2}\dot{e}_\theta^2 + \frac{\rho_\theta^2}{2\epsilon}, \label{eq.48}
\end{equation}
where $\rho_\theta\!>\!0$ indicates the maximum pitch disturbance, and the design parameter $\epsilon\!>\!0$ adjusts the trade-off between error and disturbance. 
Meanwhile, the cross terms $\frac{1}{2}\left( ({m_a L^2}/d^2) \delta_x \dot{\delta}_x \dot{e}_\theta^2 - {\lambda k_p} e_\theta \dot{e}_\theta\right)$ are eliminated by imposing the gain condition 
\begin{equation}
    \lambda > \frac{m_a L^2}{d^2} \cdot \frac{\max|\delta_x \dot{\delta}_x|}{k_p}, \label{eq.49}
\end{equation} 
which allows the Lyapunov function derivative to be upper bounded by 
\begin{equation}
    \dot{V} \leq -a \dot{e}_\theta^2 - b e_\theta^2 + \frac{\rho_\theta^2}{2\epsilon}, \label{eq.50}
\end{equation}
where $a\!=\!D_\theta\!+\!\frac{m_a g L k_d}{2d}\!+\!\frac{\lambda k_d}{2}\!-\!\frac{\epsilon}{2}$ and $b\!=\!\frac{\lambda k_p}{2}$. 
Thus, gains $k_p$, $k_d$, and $\lambda$ satisfying $D_{\theta}\!+\!\frac{m_a g L k_d}{2d}\!+\!\frac{\lambda k_d}{2}\!>\!\frac{\rho_{\theta}}{2}$ guarantee the negative definiteness of the Lyapunov function derivative $\dot{V}$. 
For disturbance-free flight ($\rho_\theta=0$), the inequality $\dot{V}\!\leq\!-a \dot{e}_\theta^2\!-\!b e_\theta^2$ holds, establishing global asymptotic stability of the pitching dynamics. 
With bounded disturbances ($\rho_\theta\neq0$), the pitching tracking system is uniformly ultimately bounded (UUB) with a convergence radius $\| e_\theta \|\!\leq\!{|\rho_\theta|}/{\sqrt{\epsilon\lambda k_p}}$, where $\epsilon$ is the decay rate coefficient in the Lyapunov analysis. 

Furthermore, real-world experiments with the \mbox{RGBlimp-Q} prototype presented in Section~\ref{Sec.experiment} validate the theoretical analysis, demonstrating stable flight and robust disturbance rejection by the proposed controller.

\subsection{A Bionic Bird's Claw for Resisting Wind}
\label{Sec.BioClaw}
\begin{figure}[htbp]
      \centering
      \includegraphics[scale=1]{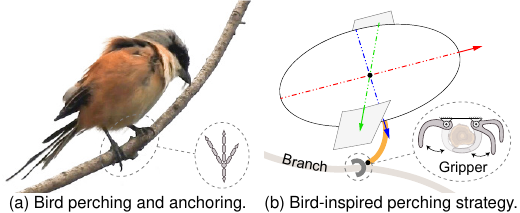}
      \caption{The gripper is a bird-inspired robotic effector mounted at the end of the continuum arm, which enables robotic blimps to catch objects for resting or resisting wind.
               (a) Birds use their claws, usually anisodactyl, to grasp branch securely and to anchor the body in a windy environment \cite{birdPic2}. 
               (b) The \mbox{RGBlimp-Q} adopts the same strategy to resist environmental wind with its gripper open for standing and closed for holding a branch. }
      \label{fig.10}
\end{figure}

With their claws, birds effortlessly grasp and perch on trees, especially in windy conditions, as illustrated in Fig.\!~\ref{fig.10}.
In contrast, current robotic blimps lack such an adaptive mechanism to withstand wind disturbances. 

Although robotic blimps feature durability, safety, and low noise levels compared to other aerial robots, they suffer from the long-standing challenge of wind disturbance, limiting their outdoor applications. 
To address this limitation, we propose a bio-inspired claw mechanism integrated with the continuum arm, mimicking behaviors of avian legs and claws. 

The bio-inspired claw is a lightweight robotic gripper mounted at the end of the continuum arm, mimicking the anisodactyl structure of a bird' foot, with the first toe pointing backwards and the remaining toes facing forward. This configuration provides a wider grasping area when the claw is open and ensures a reliable grasp when closed, thus enabling efficient aerial manipulation. 
Additionally, the buoyancy and gravity of the \mbox{RGBlimp-Q} ensure that its attitude remains stable whether the claw is open or closed. 

Furthermore, the integration of the continuum arm with the end-effector claw offers complementary benefits. Specifically, the concentrated weight at the arm's end facilitates moving mass actuation. 
The inherent flexibility and passive compliance of the continuum arm further enhance the efficiency of aerial manipulation tasks. 

\section{Prototype and Experimental Setup}
\label{Sec.prototype}
In this section, we provide a detailed description of the prototype and experimental setup. 
The moving-mass-based prototype's experimental flights are described in Section~\ref{Sec.experiment}. 

\subsection{Continuum Arm Prototype}
\begin{figure*}[p]
      \centering
      \includegraphics[scale=0.86]{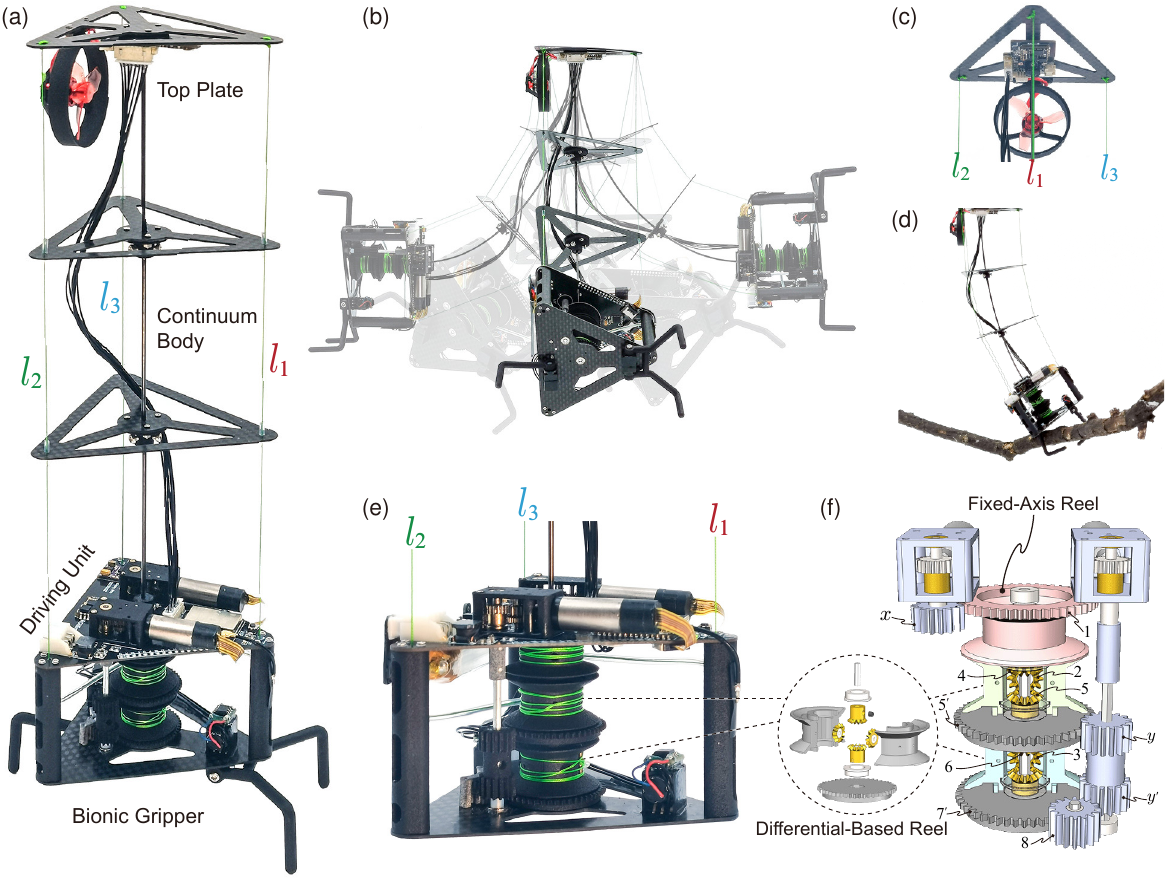}
      \vspace{-10pt}
      \caption{Prototype of the continuum arm. 
               (a) Four main parts make up the continuum arm, including the top plate, the continuum body, the driving unit, and the bionic gripper. 
               (b) The spatial motion of the continuum arm is $2$ DoF, which is constrained by the continuum body and driven by three cables.
               (c) The top plate includes a propeller and an IMU, whose circuit board and carbon fiber plate are connected to the driving unit by wires and cables, respectively. 
               (d) The bionic gripper of the prototype holds a branch for resting or resisting wind. 
               (e) The driving unit controls the continuum arm, the propeller and the bionic gripper based on sensor feedback, accounting for the main weight, i.e., $\sim \!\SI{65}{g}$ ($80\%$ total).
               (f) The gear box is composed of one fixed-axis reel, two differential-based reels, two motors with worm gear reducers, and gears of the same tooth number as designed in Section~\ref{Sec.CableDriven}.} 
      \label{fig.11}
      \vspace{20pt}
      
      \includegraphics[scale=0.85]{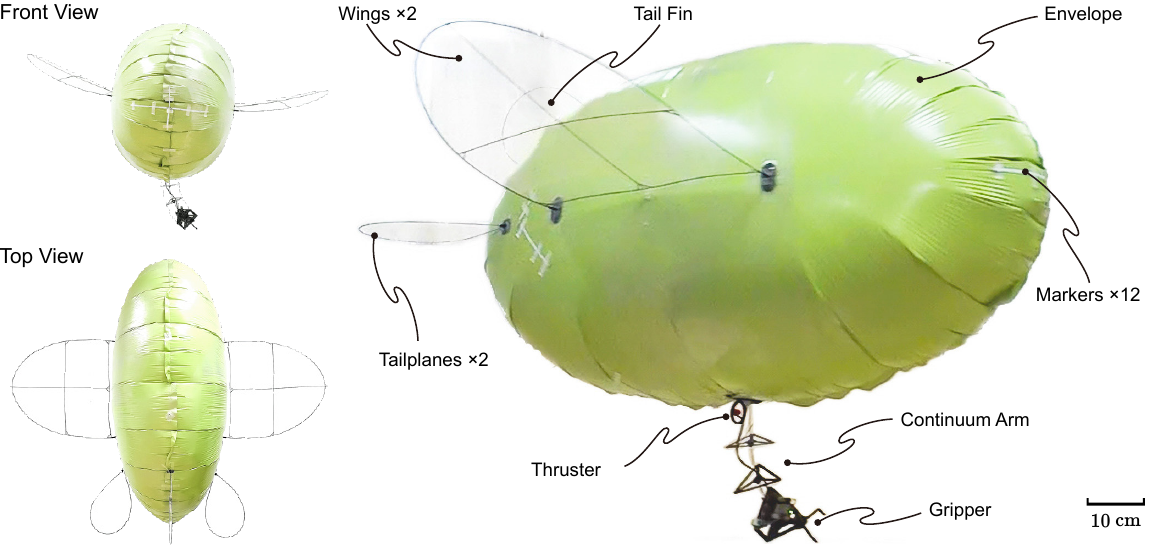}
      \vspace{-3pt}
      \caption{The \mbox{RGBlimp-Q} prototype includes an envelope, a pair of main wings, a tail fin, two tailplanes, and the continuum arm with a battery. The left parts display the prototype in front view and top view. The right part shows the components, taken during a flight trial.}
      \label{fig.12}
\end{figure*}

Following the design of the continuum arm, we have successfully developed a prototype, as shown in Fig.\!~\ref{fig.11}.
The computer-aided design (CAD) models have been made available as open source on GitHub (https://rgblimp.github.io). 
The weight of the continuum arm is $\SI{81.79}{g}$, excluding batteries, and it has a total length of $\SI{0.4}{m}$ with a backbone measuring $\SI{0.3}{m}$ made of nickel-titanium alloy. 

The continuum arm comprises four main components, arranged from top to bottom, as illustrated in Fig.\!~\ref{fig.11}(a). 
At the top plate, a propeller and an IMU are mounted, serving propulsion and sensing functions, respectively. 
This plate also acts as a connector to the envelope of \mbox{RGBlimp-Q}.
Below, the continuum body consists of three driven cables, two support plates, and the backbone. 
Meanwhile, the driven cables are constrained by the support plates to maintain a constant distance of $d\!=\!\SI{40}{mm}$ (as shown in Fig.\!~\ref{fig.6}) from the backbone while preventing twisting, thereby ensuring that the arm's motion conforms to the CC assumption. 
The driving box houses the gearbox for the decoupled cable-driven mechanism, two motors, a control unit, and another IMU. 
The decoupled mechanism is detailed in the CAD model, featuring one fixed-axis reel and two differential-based reels. 
Finally, the bionic gripper is affixed to the bottom of the continuum arm for standing or grasping objects, as illustrated in Fig.\!~\ref{fig.11}(d). 
It is worth noting that the driving cables wind in opposite directions between the fixed-axis reel and the two differential-based reels. 
Furthermore, these cables are guided from a horizontal to a vertical orientation through Teflon tubes, facilitating the movement of the continuum arm. 

The propeller, with a diameter of $\SI{31}{mm}$, is powered by a brushless DC (BLDC) motor \textit{SE0802-KV19000} and an electronic speed controller (ESC) to regulate its speed effectively.
Driving the system is a low-cost microcontroller \textit{ESP32}, leveraging the \textit{rosserial\_arduino} library to establish a \textit{Robot Operating System} (ROS) node, facilitating wireless network communication with a host computer. 

\subsection{RGBlimp-Q Prototype}
A prototype of the \mbox{RGBlimp-Q} is developed, combining the continuum arm (acting as the moving mass $m_a$) and the stationary mass $m$, encompassing an envelope, wings, and motion capture markers, as illustrated in Fig.\!~\ref{fig.12}. 
The mass distribution of the \mbox{RGBlimp-Q} prototype is detailed in Table~\ref{tab:mass}. 

\begin{table}[H]
    \caption{Mass distribution of the \mbox{RGBlimp-Q} prototype.}
    \label{tab:mass}
    \vspace*{-0.5em}
    \begin{center}
        \begin{tabular}{p{5.3cm}p{1.7cm}}
            \toprule
            Component & \hspace{2.4mm}Weight\\
            \midrule
            \textbf{Buoyant Lift}$\,^\text{a}$ $-F_b$ (Helium)          &\hspace{1.72mm}$\textbf{-194.23\,gf}$\\[1.3mm]
            \textbf{Stationary Mass}\ \,$m_0$                           &\hspace{2.65mm}$\textbf{108.69\,g}$\\[0.3mm]
            \hspace{3mm}Helium                                          &\hspace{3.73mm}$\SI{26.93}{g}$\\
            \hspace{3mm}Envelop                                         &\hspace{3.73mm}$\SI{53.74}{g}$\\
            \hspace{3mm}Main Wings                                      &\hspace{5.21mm}$\SI{8.33}{g}\times 2$\\
            \hspace{3mm}Tail Wings (Tail Fin \& Tailplanes)             &\hspace{5.21mm}$\SI{2.79}{g}\times 3$\\
            \hspace{3mm}Motion Capture Markers (Infrared LEDs)          &\hspace{5.21mm}$\SI{2.98}{g}$\\[1.3mm]
            \textbf{Moving Mass}\ \,$m_a$                               &\hspace{3.96mm}$\textbf{92.21\,g}$\\[0.3mm]
            \hspace{3mm}Battery                                         &\hspace{3.73mm}$\SI{10.42}{g}$\\
            \hspace{3mm}Continuum Arm                                   &\hspace{3.73mm}$\SI{81.79}{g}$\\[0.3mm]
            \midrule
            \textbf{Total Weight} (Net$\,^\text{b}$)                    &\hspace{5.41mm}$\textbf{6.67\,g}$\\
            \bottomrule
        \end{tabular}
    \end{center}
    \vspace{-1.0mm}
    \scriptsize{$^\text{a}$The total buoyancy from helium in gram force ($\SI{}{gf}$).} 
    \scriptsize{$^\text{b}$The total net weight stands for the weight remaining after deducting buoyancy, i.e., $m_0\!+\!m_a\!-\!F_b$.}
\end{table}

The envelope consists of an ellipsoidal mylar balloon filled with approximately $\SI{160}{}$ liters of helium, providing a total buoyant lift of $\SI{194.23}{gf}$. 
The main wings and tail wings, which include a tail fin and two tailplanes, contribute to both aerodynamic lift and attitude stability. 
The main wings, similar to those in our previous work \cite{RGBlimp}, feature a $\SI{400}{mm}$ wingspan and a $\SI{15}{deg}$ dihedral angle, adopting an aerodynamic shape. 
The tail fin is mounted vertically on the tail section of the envelope, while the pair of tailplanes is mounted horizontally.
The motion capture markers, comprising $\SI{850}{nm}$ infrared LEDs, facilitate indoor motion capture experiments. 

\subsection{Experimental Setup and System Identification}
\label{Sec.sysID}
The inertia tensor is calculated from the CAD model, yielding $\boldsymbol{J}\!=\!\mathrm{diag}(0.035,0.020,0.015)\,\SI{}{kg\!\cdot\!m^2}$. 
The gravitational constant is set to $g\!=\!\SI{9.81}{m/s^2}$. 
Utilizing the identification approach introduced in our previous work \cite{RGBlimp}, the parameters in Eq.\!~\eqref{eq.37} are identified, as listed in Table~\ref{tab:params}. 
\begin{table}[ht]
    \caption{Identified aerodynamic coefficients of the prototype. }
    \label{tab:params}
    \begin{center}
        \begin{tabular}{p{1.1cm}<{\centering}p{1cm}<{\raggedleft}p{2cm}<{\raggedleft}p{2cm}<{\raggedleft}}
            \toprule
            $C_{\hspace{-0.3mm}\Box}^\Box$\,$^*$ & \makecell[c]{$0$} & \makecell[c]{\ \ \ $\alpha$} & \makecell[c]{\ \ \ $\beta$} \\
            \midrule
            $D$   & \SI{0.243}{} & \SI{8.838}{rad^{-2}} & \SI{9.016}{rad^{-2}}\\
            $S$   & \SI{-0.082}{} & \SI{-0.285}{rad^{-2}} & \SI{-2.356}{rad^{-1}}\\
            $L$   & \SI{0.159}{} & \SI{2.938}{rad^{-1}} & \SI{8.103}{rad^{-2}}\\
            $M_1$ & \SI{-0.036}{} & \SI{0.553}{rad^{-1}} & \SI{-0.683}{rad^{-1}}\\
            $M_2$ & \SI{0.057}{} & \SI{0.093}{rad^{-1}} & \SI{5.236}{rad^{-2}}\\
            $M_3$ & \SI{0.093}{} & \SI{-0.209}{rad^{-1}} & \SI{-0.356}{rad^{-1}}\\
            \midrule
            \multicolumn{4}{l}{$K_1,K_2,K_3$\hspace{0.5cm}
            \SI{-0.073}{},\hspace{0.1cm}
            \SI{-0.052}{},\hspace{0.1cm}
            \SI{-0.032}{\ \ N\!\cdot\!m\!\cdot\!s/rad}}\\
            \bottomrule
        \end{tabular}
    \end{center}
    \vspace{-1.0mm}
    \scriptsize{$^*$ Due to space constraints, the degree of the corresponding term in the aerodynamic coefficient is inferred from its unit, e.g., $C_D^{\alpha}\alpha^2 = (\SI{8.838}{rad^{-2}})\!\cdot\!\alpha^2$. }
\end{table}

The flight performance of the \mbox{RGBlimp-Q} prototype is evaluated through numerous flight trials conducted in both indoor and outdoor settings. 
Indoor flights take place within a motion capture arena measuring $\text{L\hspace{0.3mm}}10.0\!\times\!\text{W\hspace{0.3mm}}6.0\!\times\!\text{H\hspace{0.3mm}}\SI{3.0}{m}$, equipped with ten \textit{OptiTrack} cameras. These cameras capture data at $\SI{60}{Hz}$ with an accuracy of $\SI{0.8}{mm}$ RMS in position estimation error. 
For outdoor flights, a side-view camera and an overhead camera, positioned $\SI{7}{m}$ above the ground, are used to record flight trajectories. 
Each trial, whether indoor or outdoor, utilizes an electromagnetic releaser to ensure stability and consistency in the starting pose. 
Communication between the robot and the host PC occurs via a router and ROS.

\section{Experiment and Discussion}
\label{Sec.experiment}
In this section, we present the results of flight experiments conducted with the prototype and its comparison counterparts. 
Through these experiments, we evaluate the effectiveness of the design in achieving enhanced attitude adjustment and energy efficiency. 

\subsection{Flight Maneuverability Experiment}
The proposed continuum-based moving mass actuation enables agile attitude maneuvers with less reliance on aerodynamics, as experimentally demonstrated below.  

\begin{figure}[htpb]
      \centering
      \includegraphics[scale=1]{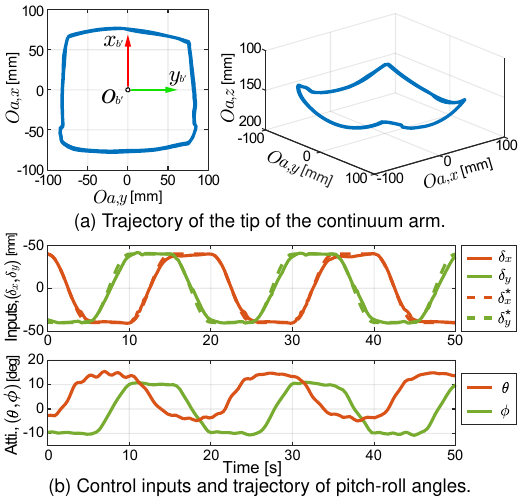} 
      \caption{Experimental results of pitch-roll adjustment through the moving mass control of the continuum arm in static floating state without propulsion. 
               (a) Square-like trajectory of the continuum arm tip (moving mass $m_a$), with its four corners tilted up. 
               (b) Trajectories of alternating sinusoidal control inputs ($\delta_x,\delta_y$) and corresponding attitude outputs pitch-roll angles ($\theta,\phi$). 
               The experimental results demonstrate the effectiveness of the moving mass control in attitude adjustment, even without aerodynamic moments and propulsion.} 
      \label{fig.13}
\end{figure}

\textit{1) Adjusting pitch-roll angles under zero propulsion}:
with the propeller shutdown and the \mbox{RGBlimp-Q} floating in the air, the continuum arm was controlled by the PID controller following predefined alternating sinusoidal control inputs, i.e., ($\delta_x^*,\delta_y^*$), as shown in Fig.\!~\ref{fig.13}(b). 
As a result, the continuum arm alternately moved the same distance along the $x_{b'}$ or $y_{'}$ directions, thus the end motion trajectory of the continuum arm formed a rectangle, as shown in Fig.\!~\ref{fig.13}(a). 

Figure~\ref{fig.13} presents pitch-roll angles of the \mbox{RGBlimp-Q} prototype adjusted by the motion of the continuum arm. 
It is observed that the inner loop control successfully tracks the desired continuum arm configuration $(\delta^*_x, \delta^*_y)$. 
Further, the input ($\delta_x,\delta_y$) independently controls the pitch and roll angles. 

\vspace{1mm}
\textit{2) Maneuvering flight with moving mass control}: 
During flight, the \mbox{RGBlimp-Q} adjusts its direction and altitude by controlling aerodynamic yaw moment and lift through roll and pitch angle adjustments.  

\begin{figure*}[t]
      \centering
      \includegraphics[scale=1]{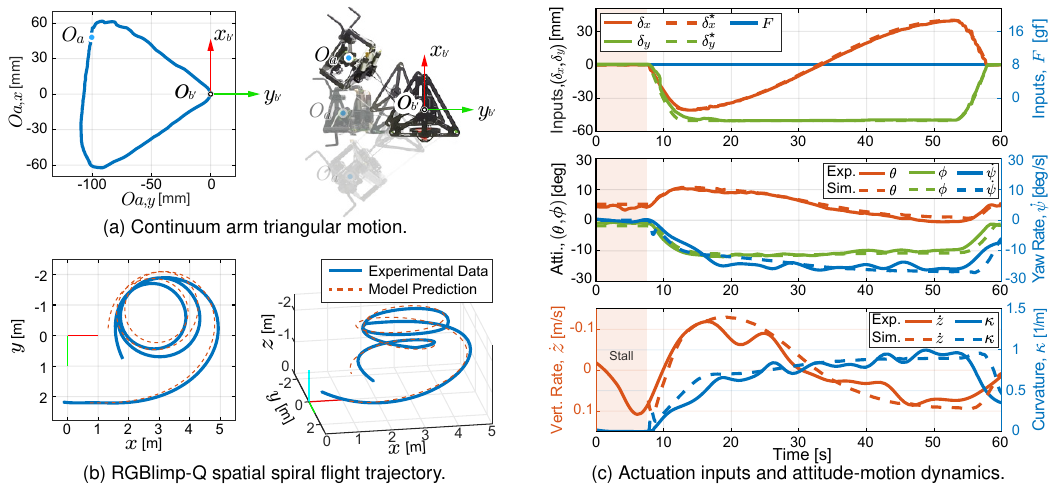}
      \caption{Experimental results and model predictions of the dynamic flight under time-varying continuum arm inputs. 
               The triangular motion of continuum-based moving mass $O_a$ results in a spiral flight trajectory with attitude-motion dynamics. 
               (a) The continuum arm, following predefined control inputs, forms a triangular trajectory. 
               (b) The flight trajectory of the robot exhibits varying curvature in the horizontal projection and varying speed in the vertical plane. 
               (c) Top: the continuum arm $(\delta_x, \delta_y)$ tracks a piecewise trigonometric function $(\delta_x^*, \delta_y^*)$.  
               Mid: pitch $\theta$ and roll $\phi$ depend on $(\delta_x, \delta_y)$, and yaw rate $\dot{\psi}$ depends on $\phi$.  
               Bottom: vertical speed $V_z$ and curvature $\kappa$ are influenced by $\theta$ and $\phi$ via aerodynamic forces.  
               Model predictions ($\dot{z}, \kappa$) for the take-off phase ($0$-$8$ seconds) are omitted because the stall AoA $\alpha$ is incompatible with the aerodynamic model. 
               }
      \label{fig.14}
\end{figure*}

Figure~\ref{fig.14} illustrates the experimental results of the maneuvering flight along with model predictions. 
The inputs, forming a triangular motion of continuum-based moving mass $O_a$ (Fig.\!~\ref{fig.14}(a)), result in a spiral flight trajectory with time-varying altitude and curvature (Fig.\!~\ref{fig.14}(b)). 
The pitch $\theta$ and roll $\phi$ are individually adjusted by the inputs $\delta_x$ and $\delta_y$, respectively. 
Despite low cruising speed about $\SI{0.5}{m/s}$, aerodynamic lift and yaw moment vary with $\theta$ and $\phi$, respectively, affecting vertical speed $V_z$ and yaw rate $\dot{\psi}$. 

\begin{figure*}[p]
      \centering
      \includegraphics[scale=1]{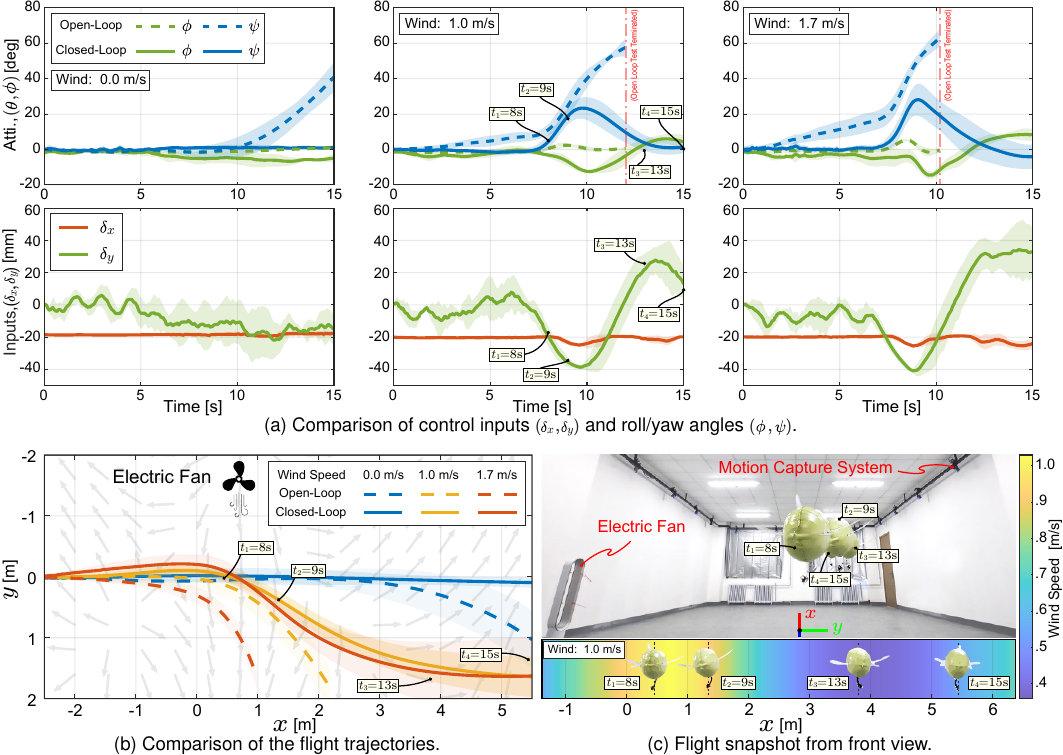} 
      \caption{Experimental results of the indoor flight trials under fan-generated wind disturbances. 
         (a) Comparison of yaw angle regulation $\psi^*\!=\!0$ under varying wind disturbances at levels $0.0, 1.0,$ and $\SI{1.7}{m/s}$: feedback control with moving mass actuation vs. open-loop flight in indoor trials, shaded areas indicating deviations. 
         The first row compares the attitude $(\phi,\psi)$ in closed-loop (solid lines) and open-loop (dashed lines) flights, with the pitch angle held constant at $\theta\!\approx\!\SI{13}{^\circ}$. 
         The second row shows inputs $(\delta_x,\delta_y)$ for closed-loop attitude control, while open-loop inputs are fixed at $(\delta_{x}^{\scriptscriptstyle\mathrm{ol}},\delta_{y}^{\scriptscriptstyle\mathrm{ol}})\!=\!(-20,0)\si{mm}$. 
         (b) Comparison of flight trajectories highlights feedback yaw control with moving mass actuation (solid lines) versus open-loop flight (dashed lines) under fan-generated wind disturbances, demonstrating robust flight attitude as the foundation for trajectory tracking. 
         Meanwhile, selected snapshot positions and CFD-simulated wind direction distribution are presented for reference.  
         (c) Wind-disturbed motion arena setup: CFD-generated speed heat maps with continuum arm dynamic response snapshots.}
      \label{fig.15}
      \vspace{1em}

      \includegraphics[scale=1]{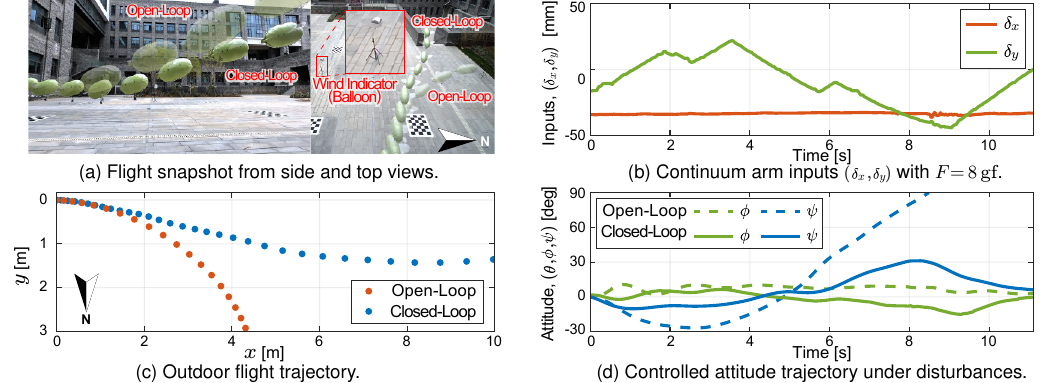} 
      \caption{Experimental results of the outdoor flight tests under natural wind disturbance. 
               (a) Closed-loop and open-loop outdoor flight trials, including side and top view snapshots and a balloon-based wind indicator.  
               (b) The trajectories of control inputs for the closed-loop trial with respect to flight time. 
               (c) Comparison of flight trajectories between open-loop and moving-mass-based yaw control under outdoor wind conditions. 
               (d) Comparison of resulting yaw angles: feedback control with moving mass actuation vs. open-loop flight in outdoor trials. 
               The outdoor experiments demonstrate that the control input of the continuum arm $\delta_y$ restores the heading angle of the robot $\psi$ following a gust disturbance, confirming the effectiveness of our moving mass control design.}
      \label{fig.16}
\end{figure*}

\subsection{Flight Robustness Experiment}
\textit{1) Flight with a robust course direction \textnormal{(}yaw angle $\psi^*\!\!=\!0$\textnormal{)}}:  
extensive indoor and outdoor trials comparing open-loop and feedback control with moving mass actuation were conducted to demonstrate the flight robustness of the \mbox{RGBlimp-Q} against wind disturbances. 
These experiments focused on regulating the yaw angle to maintain a steady course direction. 

For indoor flights, the experimental setup resembled that depicted in Fig.\!~\ref{fig.15}(c). 
An electric fan (specifically, the \textit{Dyson AM07 tower fan}) was adopted to create environmental wind disturbances at varying intensity levels. 
Three levels of wind intensity were considered, corresponding to average wind speeds of approximately $0.0$, $1.0$, and $\SI{1.7}{m/s}$, respectively. 
Each level was tested five times, with trials consisting of both open-loop flights with predefined control and closed-loop flights with feedback control. 

The comparison results of the yaw angles along with the feedback control inputs are illustrated in Fig.\!~\ref{fig.15}(a). 
It is worthy noting that as the wind level increases, the open-loop flight trajectory reached the boundaries of the flight area prematurely, resulting in significantly shorter flight time.
It is observed that the input $\delta_y$ (green lines in the second row) effectively adjusts the roll angle $\phi$ (green lines in the first row), leading to further control of the flight heading angle $\psi$ by leveraging the coupling between the aerodynamic yaw moment and the roll angle $\phi$. 
In scenarios without environmental wind disturbance, the heading angle in open-loop flights tends to drift due to inevitable asymmetries in our prototype development. The introduction of moving mass control dramatically helps mitigate this problem, ensuring flight robustness. 
When encountering environmental winds during flight, the moving mass control based on the continuum arm demonstrates a rapid response and effective disturbance rejection.

As shown in Fig.\!~\ref{fig.15}(b), the flight trajectories are compared to highlight feedback yaw control with moving mass actuation versus open-loop flight under fan-generated wind disturbances, along with CFD-simulated wind direction for reference. 
The experimental results demonstrate that a robust flight attitude is the foundation for trajectory tracking. 

For outdoor flights, a balloon was set up to indicate the direction and intensity of the natural wind, as shown in \mbox{Fig.\!~\ref{fig.16}(a)}. 
The wind speed varied between $\SI{0.5}{m/s}$ and $\SI{1.5}{m/s}$ with gusts up to $\SI{3}{m/s}$, with the wind blowing from \textit{South} to \textit{North}. 
Defining $\psi\!=\!0$ as westward with clockwise rotations positive, open-loop and closed-loop flights from \textit{East} to \textit{West} were conducted with a fixed thrust of $F\!=\!\SI{8}{gf}$, producing the trajectories shown in \mbox{Fig.\!~\ref{fig.16}(c)}. 
In closed-loop trials, the moving-mass-based feedback yaw controller regulated the yaw angle to compensate for wind disturbances. 

In Fig.\!~\ref{fig.16}(b)(d), both the input $\delta_y$ and the roll angle $\phi$ exhibit similar variations, while the heading angle $\psi$ shows an inverse trend. 
Specifically, when the heading angle $\psi$ is disturbed, the moving mass control based on the proposed continuum arm adjusts the robot's roll angle $\phi$ to counteract the disturbance with the appropriate aerodynamic yaw moment. 
The attitude of the prototype did not fully converge to the desired zero angle by the end of the flight, with the heading reaching $\SI{6}{deg}$. 
These results highlight the significant enhancement in attitude robustness achieved through the utilization of moving mass control based on the continuum arm, even in outdoor environments with complex wind disturbances. 

\begin{figure*}[thpb]
      \centering
      \includegraphics[width=0.98\textwidth]{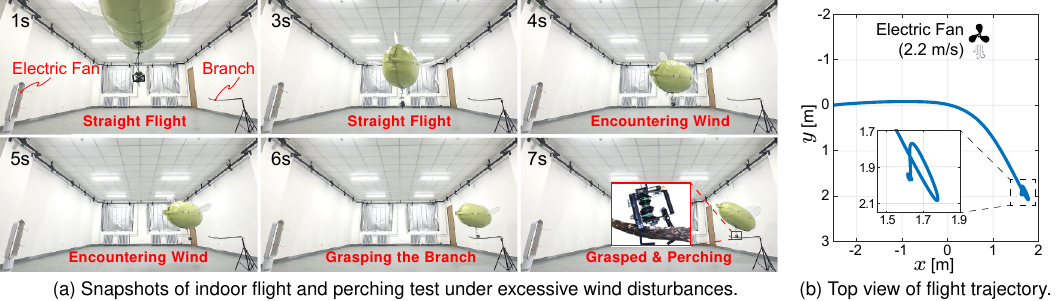}
      \caption{Experimental results of the indoor flight and perching-on-branch test under excessive wind through the use of a bio-inspired claw for resisting wind disturbances. 
               (a) The experimental snapshots illustrate the flight test process, starting from straight flight, encountering an excessive wind gust, and finally grasping a branch to resist the wind, mimicking bird behavior. 
               (b) The top view of the flight trajectory recorded by the motion capture system, with a zoomed-in view of its final phase where the robot grasps the branch and maintains position in the excessive wind. }
      \label{fig.17}
\end{figure*}

\textit{2) Grasping a branch for resisting excessive wind}: 
The effectiveness of a bio-inspired claw and associated grasping mechanism, akin to those used by birds, was demonstrated through a short flight against excessive wind ($\SI{2.2}{m/s}$) utilizing servo control based on predefined environmental priors and real-time external motion capture feedback. 
The key challenge of the perching-on-branch task is overcoming wind disturbances using the proposed bird-inspired continuum arm, enabling both aerial catching with the end-effector claw and flight attitude adjustment through moving mass actuation. 

The flight, as depicted in Fig.\!~\ref{fig.17}, unfolded in three distinct phases, i.e., flying straight ($\SI{0}{s}\,\text{-}\,\SI{3}{s}$), encountering excessive wind ($\SI{3}{s}\,\text{-}\,\SI{5}{s}$), and finally, grasping a branch to resist the wind ($\SI{5}{s}\,\text{-}\,\SI{7}{s}$). 
Corresponding feedback control strategies were implemented in each phase to counteract wind disturbances, outlined as follows: 

a) Straight flight $(\SI{0}{s}\text{-}\SI{3}{s})$: The blimp initiated from a fixed position at the release point (electromagnetic releaser) and was controlled by the proposed dual-loop PID controller to fly along the positive $x$-axis direction, that is maintaining a stable course angle $\phi=0$. 

b) Encountering strong wind $(\SI{3}{s}\text{-}\SI{5}{s})$: Upon entering the direct disturbance area of the electric fan, the heading angle was significantly deviated due to the wind disturbance. 
At this stage, the destination of the heading angle and relevant control parameters of the dual-loop PID controller were updated to facilitate rapid adjustments in both attitude and altitude, aligning the end-effector claw with the predetermined branch grasping direction.

c) Grasping a branch to resist wind $(\SI{5}{s}\text{-}\SI{7}{s})$: As the robot approached the known branch, the continuum arm was further adjusted to open the end-effector claw towards the effective grasping region based on the real-time spatial deviation. Once reaching the target, the gripper was closed to secure stable perching.  

After being exposed to excessive wind, the robot quickly veered off to the side, then adjusted its attitude to navigate towards and ultimately grasp a branch for stability. 
In the enlarged plots of the final phase (grasping) shown in Fig.\!~\ref{fig.17}(b), the robot experienced a slight swing due to inertia after successfully grasping the branch, before stabilizing its position in the windy environment. 

The perching-on-branch trials demonstrate a notable tolerance to uncertainties. 
This tolerance is facilitated by the effective graspable width of the gripper and the size of the effective grasping area of the target branch during the grasping process. 
More crucially, the moving-mass-based attitude adjustment mechanism, along with the passive compliance of the continuum arm's flexible end-effector, enhances the system's adaptability and adjustability to effectively address the grasping challenge. 

The trials consisted of five attempts: two successful grasps and stable perches, one successful grasp followed by the gripper being blown away, and two failed attempts due to excessive turbulence. 
Consequently, the success rate of the trials was approximately $\SI{40}{\%}$. 
This rate could be improved through refined control strategies and the inclusion of additional sensors, such as FPV or hand-eye cameras. 

Additionally, we acknowledge that exposed cables in the current design may interact with or become entangled in environmental obstacles (e.g., small branches), indicating the need for further analysis and design improvement.

\subsection{Comparison Against Other Actuation Mechanisms}
\begin{figure}[t]
      \centering
      \includegraphics[width=88mm]{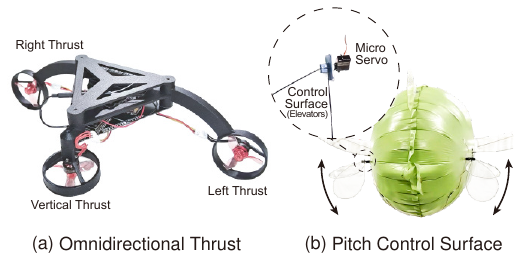}
      \caption{Comparative baselines for actuation mechanisms. 
               (a) Baseline Omni-Thrust Blimp: A reference prototype equipped with conventional omnidirectional thrusters, consisting of three BLDC propellers with electronic speed controllers (ESC), analogous to the forward thrust configuration of the \mbox{RGBlimp-Q}. 
               This baseline system shares identical aerodynamic components (wings and envelope) and overall net weight with \mbox{RGBlimp-Q}, ensuring a fair comparison. 
               (b) Pitch Control Surface: A baseline control-surface-based counterpart is configured by replacing the fixed tailplanes with elevators, allowing deflection within $\delta_e \in [-45^\circ, 45^\circ]$. While preserving the same overall configuration, this prototype utilizes aerodynamic moment generated by the elevator for pitch adjustment, actuated via micro servos. }
      \label{fig.18}
\end{figure}

The primary distinction of the proposed \mbox{RGBlimp-Q} lies in its novel actuation mechanism---continuum-based \mbox{moving} mass actuation for flight attitude adjustment. 
This section conducts experimental comparisons and analyses of the \mbox{RGBlimp-Q} prototype and counterpart prototypes of the same aerodynamic and buoyant configuration but differing in actuation mechanisms: omnidirectional thrusters or control surfaces as shown in Fig.\!~\ref{fig.18}. 
To isolate the performance differences attributable solely to the actuation mechanisms, all flight experiments were conducted under standardized conditions. 
Specifically, all prototypes employed identical envelope and wing configurations (Fig.\!~\ref{fig.12}) and were tested in a consistent experimental environment, maintaining similar wind conditions, ambient temperature, and air pressure. 
Furthermore, the total net weight of all prototypes was precisely matched to $m\!=\!\SI{6.67}{g}\pm\SI{0.30}{g}$ using onboard ballast, as per Table~\ref{tab:mass}. 

\vspace{1mm}
\textit{1) Moving Mass\ \,vs. Omnidirectional Thrust---Attitude \mbox{adjustment} capability}:\quad
In alignment with the experimental setup employed in the indoor flight trials (Fig.\!~\ref{fig.15}(c)), comparative trials were conducted to assess the pitch-roll attitude adjustment capability of the \mbox{RGBlimp-Q} (moving mass actuation) and the conventional omnidirectional thrust actuation (Omni-Thrust). 
These trials were performed under identical wind disturbance conditions, specifically a crosswind of approximately $\SI{1.0}{m/s}$, generated by a side-mounted electric fan. 
Both prototypes used a PID feedback controller for pitch-roll stabilization to assess performance differences in attitude regulation during flight. 

\begin{figure}[t]
      \centering
      \includegraphics[scale=1]{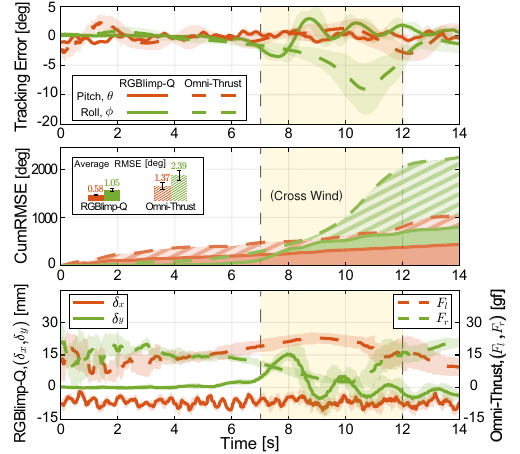}
      \caption{The comparative results of flight with robust pitch-roll attitude, where the desired pitch angles are $\theta_\mathrm{R}^*=\SI{10}{^\circ}$ for \mbox{RGBlimp-Q} and $\theta_\mathrm{O}^*=0$ for the Omni-Thrust prototype, and the desired roll angles are $\phi_\mathrm{R}^*=\phi_\mathrm{O}^*=0$.} 
      \label{fig.19}
\end{figure}

The upper panel of Fig.\!~\ref{fig.19} presents the cumulative root mean squared error (CumRMSE) of pitch-roll attitude, calculated as the sum of RMSE over time for both actuation mechanisms. 
The lower panel illustrates the corresponding inputs for \mbox{RGBlimp-Q} and Omni-Thrust, specifically the decoupled cable-driven continuum inputs $(\delta_x, \delta_y)$ and the left-right omnidirectional thrust $(F_l, F_r)$, respectively. 
The results show that within the crosswind disturbance ($7$-$12$ seconds), the roll CumRMSE for Omni-Thrust (dashed green line) increases sharply, whereas \mbox{RGBlimp-Q} (solid green line) shows a more gradual rise. 
Similarly, the pitch CumRMSE for \mbox{RGBlimp-Q} (solid red line) increases slowly, while Omni-Thrust (dashed red line) exhibits a significant rise. 

These trends demonstrate that while both systems are affected by crosswinds, the moving mass actuation in \mbox{RGBlimp-Q} mitigates long-term error accumulation in pitch and roll, ensuring better pitch-roll stability. 
The CumRMSE metric effectively highlights the advantages of moving mass actuation for robust attitude performance. 

\vspace{1mm}
\textit{2) Moving Mass\ \  vs. Omnidirectional Thrust---Flight energy efficiency}:\quad 
To ensure a fair evaluation of the performance differences in flight energy efficiency between the actuation mechanisms, both prototypes are configured with identical propeller(s) and ESC(s) for thrust generation and are powered by the same driving unit (ESP32 SoC) and battery ($\SI{7.4}{v}$, $\SI{150}{mAh}$, Li-Po). 
The primary difference in energy consumption arises from the actuation strategy. 
The proposed continuum-based moving mass actuation employs a single forward thrust and two actuators for pitch-roll attitude control. In contrast, the conventional omnidirectional thrust mechanism requires at least three thrust vectors---namely, left and right forward thrusts and a vertical thrust---resulting in increased energy expenditure. 

\begin{figure}[htbp]
      \centering
      \,\includegraphics[scale=1]{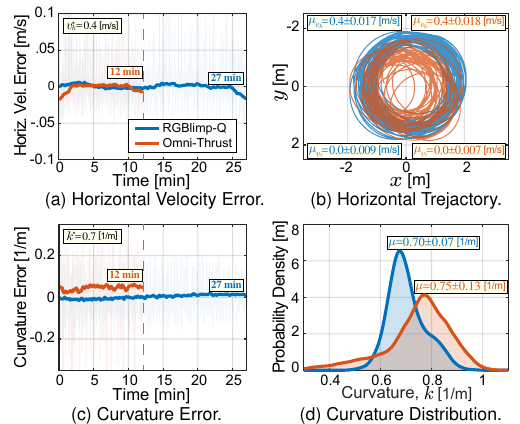}
      \caption{Comparative results of flight endurance in terms of duration and distance between \mbox{RGBlimp-Q} and the Omni-Thrust prototype.} 
      \label{fig.20}
\end{figure}

The endurance performance comparison, evaluated in terms of flight time and distance, was conducted through continuous circling flights maintained at a consistent target altitude, velocity, and curvature until the battery could no longer support sustained flight. 
In each flight trial, the trajectory was recorded by the motion capture system. 
The corresponding results are presented in Fig.\!~\ref{fig.20}, with quantitative comparison metrics listed in Table~\ref{tab:endurance}. 

The comparison between the proposed \mbox{RGBlimp-Q} and the conventional omnidirectional blimp reveals significant performance improvements, particularly in energy efficiency, as measured by flight time and distance. 
At a comparable slow cruising speed of approximately $\SI{0.4}{m/s}$ (Fig.\!~\ref{fig.20}(a)), the proposed system exhibits nearly a twofold improvement in both flight duration and traveled distance. 
These enhancements are attributed to significantly lower power consumption ($\SI{5.6}{mAh/min}$ vs. $\SI{12.5}{mAh/min}$), and a reduced energy-to-distance ratio ($\SI{1.8}{mWh/m}$ vs. $\SI{4.0}{mWh/m}$). 

\vspace*{-1em}
\noindent
\begin{table}[ht]
    \caption{Flight Endurance Performance: Quantitative Comparison between the \mbox{RGBlimp-Q} and Omni-Thrust Prototype.}
    \label{tab:endurance}
    \vspace{-2mm}
    \begin{center}
        \begin{tabular}{p{2.5cm}<{}p{1.35cm}<{}p{1.4cm}<{\centering}p{1.5cm}<{\centering}}
            \toprule
            \hspace{4.5em}Metrics & & RGBlimp-Q &  Omni-Thrust \\
            \midrule
            \hspace{0.1em}Time Duration        &[\SI{}{min}]       & \underline{$27$}      & $12$\\[0.8mm]
            \hspace{0.1em}Trajectory Length    &[\SI{}{m}]         & \underline{$605$}     & $274$\\[0.8mm]
            \hspace{0.1em}Horizontal Vel. STD  &[\SI{}{m/s}]       & \underline{$0.017$}   & $0.018$\\[0.8mm]
            \hspace{0.1em}Vertical Vel. STD    &[\SI{}{m/s}]       & $0.009$               & \underline{$0.007$}\\[0.8mm]
            \hspace{0.1em}Curvature STD        &[\SI{}{1/m}]       & \underline{$0.07$}    & $0.13$\\[0.8mm]
            \hspace{0.1em}Power Rate           &[\SI{}{mAh/\!min}] & \underline{$5.6$}     & $12.5$\\[0.8mm]
            \hspace{0.1em}Specific Energy      &[\SI{}{mWh/\!m}]   & \underline{$1.8$}     & $4.0$\\
            \bottomrule
        \end{tabular}
    \end{center}
\end{table}
\vspace*{-0.5em}

As illustrated in Table~\ref{tab:endurance}, the energy efficiency is demonstrated by a $\SI{125}{\%}$ increase in flight duration ($\SI{27}{min}$ vs. $\SI{12}{min}$), resulting from the moving mass actuation mechanism that effectively minimizes thrust loss by leveraging gravitational and aerodynamic moments. 
The \mbox{RGBlimp-Q} prototype also achieves a $\SI{2.21}{}$ longer flight path ($\SI{605}{m}$ vs. $\SI{274}{m}$) compared to the Omni-Thrust, while  maintaining a more stable path curvature $\sigma_{\kappa}^\mathrm{R}<\sigma_{\kappa}^\mathrm{O}$. 
Curvature distribution analysis (Figs.\!~\ref{fig.20}(b)(c)(d)) shows a $\SI{46}{\%}$ reduction in standard deviation ($\SI{0.07}{m^{-1}}$ vs. $\SI{0.13}{m^{-1}}$), confirming the smoother trajectory enabled by the novel actuation mechanism. 

These results highlight the significant enhancement in energy efficiency enabled by the proposed moving mass actuation mechanism in robotic blimps. 

\begin{figure}[htpb]
      \centering
      \includegraphics[scale=1]{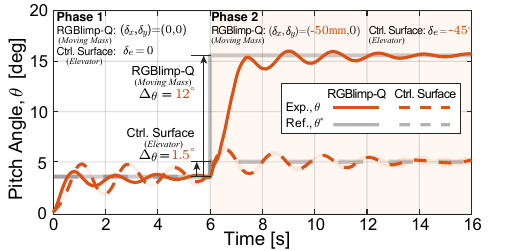}
      \caption{Comparison results of pitch adjustment at low cruising speed using moving mass and control surface mechanisms. Pitch adjustment performance is evaluated under low-speed cruising conditions ($\|\boldsymbol{v}\|\!=\!\SI{0.8}{m/s}$) in a no-wind environment.}
      \label{fig.21}
\end{figure}

\vspace{1mm}
\textit{3) Moving Mass\!~~vs.~Control Surface---Flight attitude adjustment during low-speed flight}:\quad 
Flight trials were conducted using prototypes with moving mass (\mbox{RGBlimp-Q}) and control surface actuation (Fig.\!~\ref{fig.18}(b)) under a fixed forward thrust of $\SI{7}{gf}$. 
Step changes were applied to pitch-related control inputs: $\delta_x$ for moving mass actuation and $\delta_e$ for elevator deflection. 
In the initial phase of each trial, all control inputs were set to zero, i.e., $(\delta_x,\delta_y)\!=\!0$ and $\delta_e\!=\!0$. 
At $t\!=\!\SI{6}{s}$, the pitch-related inputs were step-changed to $\delta_x\!=\!\SI{-50}{mm}$ and $\delta_e\!=\!\SI{-45}{\deg}$. 
Each configuration was tested over 10 repeated trials in an indoor motion environment with no wind disturbances, and the resulting pitch angle responses are presented in Fig.\!~\ref{fig.21}. 

The experimental results clearly demonstrate the superior performance of the moving mass actuation in regulating the pitch attitude of low-speed robotic blimps. 
Under identical control inputs, the moving mass prototype achieved a pitch angle change of approximately $\SI{12}{\deg}$, in contrast to the $\SI{1.5}{\deg}$ change observed with the elevator-based system. 
This discrepancy is primarily attributed to the reduced effectiveness of aerodynamic moments at low flight speeds, which limits the authority of conventional control surfaces. 
Conversely, the moving mass mechanism leverages gravitational moments to induce substantial attitude changes, thereby enhancing control authority under low-speed conditions. 
These findings validate the effectiveness of the proposed moving mass actuation strategy in enabling agile and robust pitch control in robotic blimps operating in low-speed regimes. 

\subsection{Continuum Arm Reliability and Repeatability}
\begin{figure}[htpb]
      \centering
      \includegraphics[width=88mm]{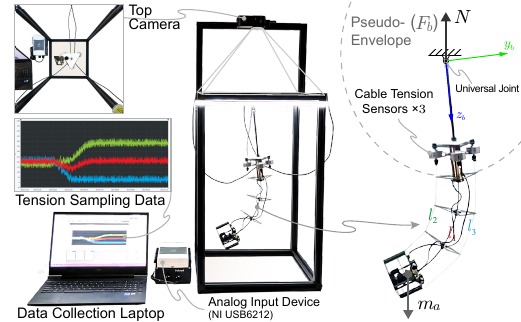}
      \caption{Experimental setup for evaluating the reliability and repeatability of the continuum arm.}
      \label{fig.22}
\end{figure}

To evaluate the working duration and tendon reliability of the cable-driven continuum mechanism, we conducted a cyclic motion test using a delicately designed experimental apparatus, as shown in Fig.\!~\ref{fig.22}. 
The platform comprises a continuum arm prototype identical in design, affixed to a lightweight carbon fiber rod mounted via a $2$-DoF universal joint to a structural frame. 
This joint-link configuration emulates a pseudo buoyant envelope, where the support force $N$ simulates buoyancy $F_b$, and the link length $h$ represents the vertical distance from the pseudo-envelope's center of buoyancy (CB) to the base frame $O_{b'}\text{-}x_{b'}y_{b'}z_{b'}$ of the continuum arm, replicating the operational conditions of moving mass actuation, as illustrated in Fig.\!~\ref{fig.4}. 
Three actuation cables equipped with integrated tension sensors are continuously monitored via an NI USB-6212 analog input module to evaluate the mechanical load. 
Two IMUs are placed at both ends of the arm to estimate the system's attitude via an Extended Kalman Filter (EKF), from which the arm's configuration is computed in $q$-parametrization $(\delta_x, \delta_y)$. 
This configuration feeds into an inner-loop feedback controller to track time-varying reference inputs $(\delta_x^*(t), \delta_y^*(t))$ during repeated cycle motions. 

\begin{figure}[htpb]
      \centering
      \includegraphics[scale=1]{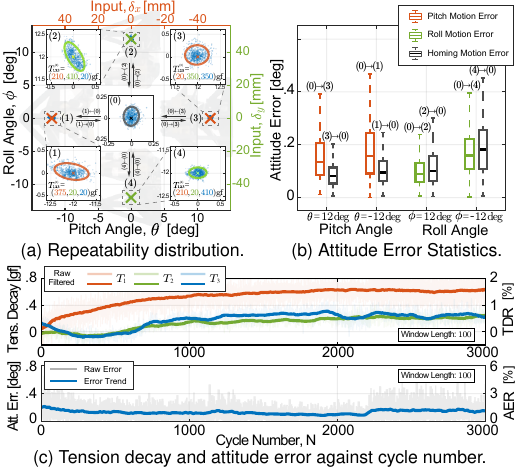}
      \caption{Performance evaluation of the cable-driven continuum arm in terms of reliability and repeatability. 
        (a) End-effector pitch-roll repeatability during repeated cycle motions between $\theta,\phi = \pm\SI{12}{\deg}$ with homing motions, visualized using scatter plots and $2\sigma$ confidence ellipses; yielding mean standard deviations $\bar{\sigma}_\theta\!=\!\SI{0.11}{\deg}$, $\bar{\sigma}_\phi\!=\!\SI{0.10}{\deg}$. 
        (b) Box plot of attitude errors across all trials, demonstrating tight variation bounds ($\leq\SI{0.3}{\deg}$), further validating high repeatability.
        (c) Reliability over $3000$ actuation cycles, showing tension decay and attitude error trends via moving mean (window=$100$), confirming consistent performance.} 
      \label{fig.23}
\end{figure}

The test protocol involved sequential round-trip actuation from $q_{(0)}^*\!\!=\!\!(0,0)$ to four directional configurations: front $q_{(1)}^*\!\!=\!\!(+50,0)\si{mm}$, right $q_{(2)}^*\!\!=\!\!(0,+50)\si{mm}$, back $q_{(3)}^*\!\!\!=\!\!(-50,0)\si{mm}$, and left $q_{(4)}^*\!\!=\!\!(0,-50)\si{mm}$, as illustrated in Fig.\!~\ref{fig.23}. 
This motion sequence was repeated for over $3000$ round-trip cycles, covering all four directional axes, with a cumulative operation duration exceeding $\SI{18}{hours}$. 

Figure~\ref{fig.23}(a) illustrates pitch-roll repeatability during repeated cycle motions by plotting $(\theta,\phi)$ and corresponding inputs $(\delta^*_x,\delta^*_y)$ using zoomed-in scatter points and $2\sigma$ confidence ellipses for five configurations $q_{(k)}^*, (k\!\!=\!\!0,1,2,3,4)$. 
The corresponding attitude of the pseudo-envelope: the pitch angles $\theta\!=\!\mp\SI{12}{\deg}$ for $q_{(1)}^*,q_{(3)}^*$, the roll angle $\phi\!=\!\pm\SI{12}{\deg}$ for $q_{(2)}^*,q_{(4)}^*$, and the initial attitude $\theta\!=\!\phi\!=\!0$ for $q_{(0)}^*$. 
The repeatability test shows that $(\theta,\phi)$ has pitch-roll mean standard deviations of $\bar{\sigma}_\theta\!=\!\SI{0.11}{\deg}$ and $\bar{\sigma}_\phi\!=\!\SI{0.10}{\deg}$. 
Additionally, the average tension $T_i^{(k)}\, (i\!=\!1,2,3; k\!=\!1,2,3,4)$ for cables $i$ at $q_{(k)}^*$ is provided. 

Box-plot analysis in Fig.\!~\ref{fig.23}(b) further confirms the narrow distribution of attitude errors, consistently within $\leq\SI{0.3}{\deg}$.
These results validate the robustness of the proposed cable-driven design with the decoupled mechanism against tendon slacking and breakage. 
The system consistently achieves high repeatability in positioning, even under prolonged operation over extended durations and cycle counts. 
Overall, the continuum arm mechanism meets the stringent requirements for sustained, high-cycle performance necessary for moving-mass actuation in robotic gliding blimps. 

As shown in Fig.\!~\ref{fig.23}(c), system reliability is evaluated based on cable tension degradation and attitude error trends. 
Cable $l_1$ showed a tension decay of $\SI{0.6}{gf}\,(\SI{1.8}{\%})$, while cables $l_2$ and $l_3$ exhibited $\SI{0.3}{gf}\,(\SI{0.8}{\%})$ reduction relative to initial values. 
Attitude errors remained within $\SI{0.2}{\deg}\,(\SI{1}{\%})$ for both pitch and roll throughout the test duration.

\section{Conclusion}
\label{Sec.conclusion}
In this article, we presented a comprehensive design of a bird-inspired robotic gliding blimp, termed \mbox{RGBlimp-Q}, engineered to exhibit enhanced maneuverability and resilience through the integration of an active continuum arm. 
This flying robot leverages the continuum arm for precise moving mass control, facilitating agile flight attitude maneuvers. 
In addition, the inclusion of a bionic claw at the end of the arm empowers the robot to grasp objects, akin to bird behavior, enabling it to withstand wind disturbances effectively. 
Extensive indoor and outdoor flight experiments were conducted, the results of which validated the effectiveness of the proposed design. 

In future work, we will investigate more advanced control strategies such as model predictive control (MPC) and learning-based control to fully exploit the capabilities of the \mbox{RGBlimp-Q}. 
Autonomous perching and aerial manipulation, using onboard perception such as FPV or hand-eye cameras, will be developed through successive prototypes. 
Additionally, substantial optimization potential remains in the weight-strength trade-off. 
By selecting an optimized elasticity modulus for the continuum backbone, we can reduce the excessive strength margin, allowing low-density materials, such as foam-based composites, to effectively decrease both payload and buoyancy requirements. 
In addition, we will explore the integration of aerodynamic and buoyant lift mechanisms within a unified aerodynamically shaped envelope (e.g., a hybrid aerodynamic-buoyant flying wing), eliminating the need for additional wings. This approach will be investigated in subsequent iterations of the work. 

In application, we envision deploying the \mbox{RGBlimp-Q} for real-world applications such as indoor or outdoor environmental monitoring and mapping, leveraging its enhanced durability and human-robot-interaction (HRI) safety to address diverse challenges. 
Moreover, it is worthwhile to explore aerial robotic manipulators that utilize the proposed bird-inspired continuum arm, applicable to both robotic blimps and rotary-wing aerial robots. 
Last but not least, robotic gliding blimps hold promise as an aerial experimental platform for simulating and rapidly validating the design of underwater gliders, leveraging the similarity in dynamic mechanisms that combine aero-/hydro-dynamic lift with static buoyant lift. 

\section*{Acknowledgments}
The authors would like to thank Prof. Zhongkui Li for his help in the motion capture experiment, and Zeyu Sha for help with the CFD simulation. 
The authors also gratefully acknowledge the reviewers and editors for their constructive comments and suggestions that greatly improved the quality of the manuscript. 

{
\appendix[]
\subsection*{\mbox{A.\ Proof of Invertibility of Matrix $\boldsymbol{A}$ in Eq.\!~\eqref{eq.34}}}
\label{Appendix.A}
\setcounter{equation}{0}
\renewcommand{\theequation}{A.\arabic{equation}} 
We prove that the allocation matrix $\boldsymbol{A}$ is invertible by analyzing the block matrix $\boldsymbol{M}$ defined as:
\begin{equation}
    \boldsymbol{M} = 
    \begin{bmatrix}
        (m_0 + m_a)\boldsymbol{I} & -\boldsymbol{R}\boldsymbol{l}_g^{\times} \\
        \boldsymbol{l}_g^{\times}\boldsymbol{R}^\mathrm{T} & \boldsymbol{J} - m_a (\boldsymbol{r}_a^{\times})^2
    \end{bmatrix},
    \label{eq:block_matrix}
\end{equation}
where $\boldsymbol{A} = \boldsymbol{M}^{-1}$. 
The invertibility of $\boldsymbol{A}$ is established through the application of the Schur complement and positive-definiteness properties.

\begin{lemma}
\label{lemma:D_invertible}
The lower block $\boldsymbol{J}_\mathrm{eff} = \boldsymbol{J} - m_a (\boldsymbol{r}_a^{\times})^2$ is positive definite and invertible.
\end{lemma}

\vspace*{-1em}
\begin{proof}
First, the positive definiteness of the inertia matrix $\boldsymbol{J}$ for the stationary mass $m_0$ is inherent in its definition, i.e., 
\begin{equation}
    \boldsymbol{J} = \int_{m_0} \left( \|\boldsymbol{r}\|^2 \boldsymbol{I} - \boldsymbol{r}\boldsymbol{r}^\mathrm{T} \right) \mathrm{d}m,
    \label{eq:moment}
\end{equation}
where $\boldsymbol{r}$ is the distance vector of $\mathrm{d}m$ with respect to the origin of the body-fixed frame. 
The operator $(\boldsymbol{r}_a^{\times})^2$ is given by
\begin{equation}
    (\boldsymbol{r}_a^{\times})^2 = -\|\boldsymbol{r}_a\|^2 \boldsymbol{I} + \boldsymbol{r}_a \boldsymbol{r}_a^\mathrm{T},
    \label{eq:skew_square}
\end{equation}
which is symmetric. 
For any non-zero vector $\boldsymbol{x}\!\in\!\mathbb{R}^3$, we have
\begin{align}
    \boldsymbol{x}^\mathrm{T}(\boldsymbol{r}_a^{\times})^2\boldsymbol{x} &= -\left \| \boldsymbol{r}_a \right \|^2 \left \| \boldsymbol{x} \right \|^2 + \left ( \boldsymbol{x}^\mathrm{T}\boldsymbol{r}_a\right )^2.\nonumber\\[-0mm]
    &\leq -\left \| \boldsymbol{r}_a \right \|^2 \left \| \boldsymbol{x} \right \|^2 + \left \| \boldsymbol{x} \right \|^2\left \| \boldsymbol{r}_a \right \|^2  = 0.
\end{align}
Thus, $(\boldsymbol{r}_a^{\times})^2$ is negative semi-definite. 
Consequently, the term $m_a (\boldsymbol{r}_a^{\times})^2$ is also negative semi-definite, as $m_a\!>\!0$. 
Therefore, $\boldsymbol{J}_\mathrm{eff} = \boldsymbol{J} - m_a (\boldsymbol{r}_a^{\times})^2$ represents the sum of a positive definite matrix and a positive semi-definite matrix. This guarantees that $\boldsymbol{J}_\mathrm{eff}$ is positive definite. 
Consequently, we have $\det(\boldsymbol{J}_\mathrm{eff}) > 0$, confirming that $\boldsymbol{J}_\mathrm{eff}$ is invertible. 
\end{proof}

\begin{lemma}
\label{lemma:schur}
The Schur complement of $\boldsymbol{J}_\mathrm{eff}$ in $\boldsymbol{M}$ is positive definite. 
\end{lemma}

\begin{proof}
The Schur complement $\boldsymbol{S}$ of $\boldsymbol{J}_\mathrm{eff}$ is defined as 
\begin{equation}
    \boldsymbol{S} = (m_0 + m_a)\boldsymbol{I} + \boldsymbol{R}\boldsymbol{l}_g^{\times} \boldsymbol{J}_\mathrm{eff}^{-1} \boldsymbol{l}_g^{\times}\boldsymbol{R}^\mathrm{T}.
    \label{eq:schur_complement}
\end{equation}
The first term $(m_0 + m_a)\boldsymbol{I}$ is positive definite. 
The second term $\boldsymbol{R}\boldsymbol{l}_g^{\times} \boldsymbol{J}_\mathrm{eff}^{-1} \boldsymbol{l}_g^{\times}\boldsymbol{R}^\mathrm{T}$ is symmetric positive semi-definite, as $\boldsymbol{J}_\mathrm{eff}^{-1}$ is positive definite. 
Therefore, we conclude that $\boldsymbol{S}$ is positive definite, and $\det(\boldsymbol{S}) > 0$. 
\end{proof}

\begin{theorem}
The matrix $\boldsymbol{M}$ in Eq.\!~\eqref{eq:block_matrix} is invertible, and therefore the allocation matrix $\boldsymbol{A} = \boldsymbol{M}^{-1}$ exists. 
\end{theorem}

\begin{proof}
Using the Schur complement formula, the determinant of $\boldsymbol{M}$ is given by 
\begin{equation}
    \det(\boldsymbol{M}) = \det(\boldsymbol{J}_\mathrm{eff}) \det(\boldsymbol{S}).
    \label{eq:determinant}
\end{equation}
From Lemmas~\ref{lemma:D_invertible} and \ref{lemma:schur}, we know that $\det(\boldsymbol{J}_\mathrm{eff}) > 0$ and $\det(\boldsymbol{S}) > 0$. 
Therefore, it follows that $\det(\boldsymbol{M}) > 0$, establishing that  $\boldsymbol{M}$ is invertible. 
Consequently, the allocation matrix $\boldsymbol{A} = \boldsymbol{M}^{-1}$ is well-defined. 
\end{proof}
}


\bibliographystyle{IEEEtran}
\bibliography{Reference}
 
\vspace{11pt}

\begin{IEEEbiography}[{\includegraphics[width=1in,height=1.25in,clip,keepaspectratio]{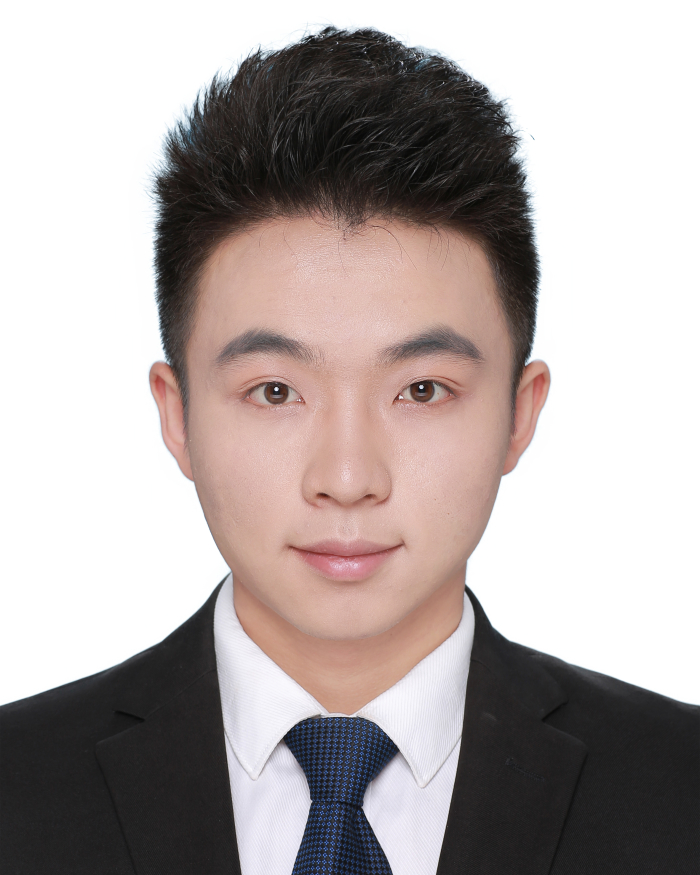}}]{Hao Cheng}
received the Bachelor's degree in New Energy Science and Engineering (Wind Power) from North China Electric Power University, China, in 2017, and the Master's degree in Control Engineering from Tsinghua University, China, in 2021. 
He is currently working toward the Ph.D. degree in Mechanics at Peking University, China, under the supervision of Prof. F. Zhang. 
His research interests include Light-Than-Air Aerial Vehicles, Bioinspired Robotics, and Continuum Robots.
\end{IEEEbiography}

\vspace{11pt}

\begin{IEEEbiography}[{\includegraphics[width=1in,height=1.25in,clip,keepaspectratio]{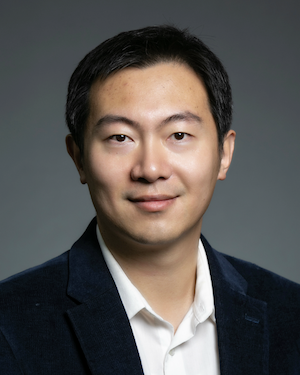}}]{Feitian Zhang}
received the Bachelor's and Master's degrees in Automatic Control from Harbin Institute of Technology, Harbin, China, in 2007 and 2009, respectively, and the Ph.D. degree in Electrical and Computer Engineering from Michigan State University, East Lansing, MI, in 2014. 

Feitian Zhang is currently an Associate Professor in the School of Advanced Manufacturing and Robotics at Peking University. 
Prior to joining Peking University, he was an Assistant Professor in the Department of Electrical and Computer Engineering at George Mason University (GMU), Fairfax, VA, and the founding director of the Bioinspired Robotics and Intelligent Control Laboratory (BRICLab) from 2016 to 2021. 
His research interests include Bioinspired Robotics, Control Systems, Artificial Intelligence, Underwater Vehicles, and Aerial Vehicles. 
\end{IEEEbiography}

\vfill

\end{document}